\documentclass[final]{cvpr}

\usepackage{times}
\usepackage{graphicx}
\usepackage{amsmath}
\usepackage{amssymb}
\usepackage{graphicx}
\usepackage{amsmath}
\usepackage{amssymb}
\usepackage{algorithm}
\usepackage[noend]{algpseudocode}
\usepackage{booktabs}

\usepackage[pagebackref=true,breaklinks=true,colorlinks,bookmarks=false]{hyperref}

\begin{document}

\title{Contextually Plausible and Diverse 3D Human Motion Prediction}

\author{
Sadegh Aliakbarian$^{1,2}$, Fatemeh Saleh$^{1,2}$, Lars Petersson$^{3}$, Stephen Gould$^{1,2}$, Mathieu Salzmann$^{4}$\\
$^{1}$ Australian National University, $^{2}$ ACRV, $^{3}$ Data61-CSIRO, $^{4}$ CVLab-EPFL
}

\maketitle
%\thispagestyle{empty}

%%%%%%%%% ABSTRACT
\begin{abstract}
   We tackle the task of diverse 3D human motion prediction, that is, forecasting multiple plausible future 3D poses given a sequence of observed 3D poses. In this context, a popular approach consists of using a Conditional Variational Autoencoder (CVAE). However, existing approaches that do so either fail to capture the diversity in human motion, or generate diverse but semantically implausible continuations of the observed motion. In this paper, we address both of these problems by developing a new variational framework that accounts for both diversity and context of the generated future motion. To this end, and in contrast to existing approaches, we condition the sampling of the latent variable that acts as source of diversity on the representation of the past observation, thus encouraging it to carry relevant information. Our experiments demonstrate that our approach yields motions not only of higher quality while retaining diversity, but also that preserve the contextual information contained in the observed 3D pose sequence. 
\end{abstract}

%%%%%%%%% BODY TEXT
\section{Introduction}
Human motion prediction is the task of forecasting plausible 3D human motion continuation(s) given a sequence of past 3D human poses. To address this problem, prior work mostly relies on recurrent encoder-decoder architectures, where the encoder processes the observed motion, and the decoder generates a single 
future trajectory given the encoded representation of the past~\cite{barsoum2018hp,gui2018adversarial,kundu2018bihmp,martinez2017human,pavllo2019modeling,pavllo2018quaternet,walker2017pose,wei2019motion}. While this approach yields valid future motion, it tends to ignore the fact that human motion is stochastic in nature; given one single observation, multiple diverse continuations of the motion are likely and plausible. The lack of stochasticity of these encoder-decoder methods ensues from the fact that both the network operations and the sequences in the training dataset are deterministic\footnote{For complicated tasks such as motion prediction, the training data is typically insufficiently sampled, in that, for any given condition, the dataset contains only a single sample, in effect making the data appear deterministic. For instance, in motion prediction, we never observed twice the same past motion with two different future ones.}.
In this paper, we introduce an approach to modeling this stochasticity by learning \emph{multiple modes} of human motion. We focus on generating both diverse \textit{and} contextually and semantically plausible motion predictions. By contextually plausible, we mean motions that are natural continuations of an observed sequence of 3D poses, preserving and continuing the context depicted by the observation. For instance, when the observed poses depict a person walking, we expect the network to predominantly predict future motions corresponding to different walking modes.

\begin{figure}
    \centering
    \includegraphics[width=0.47\textwidth]{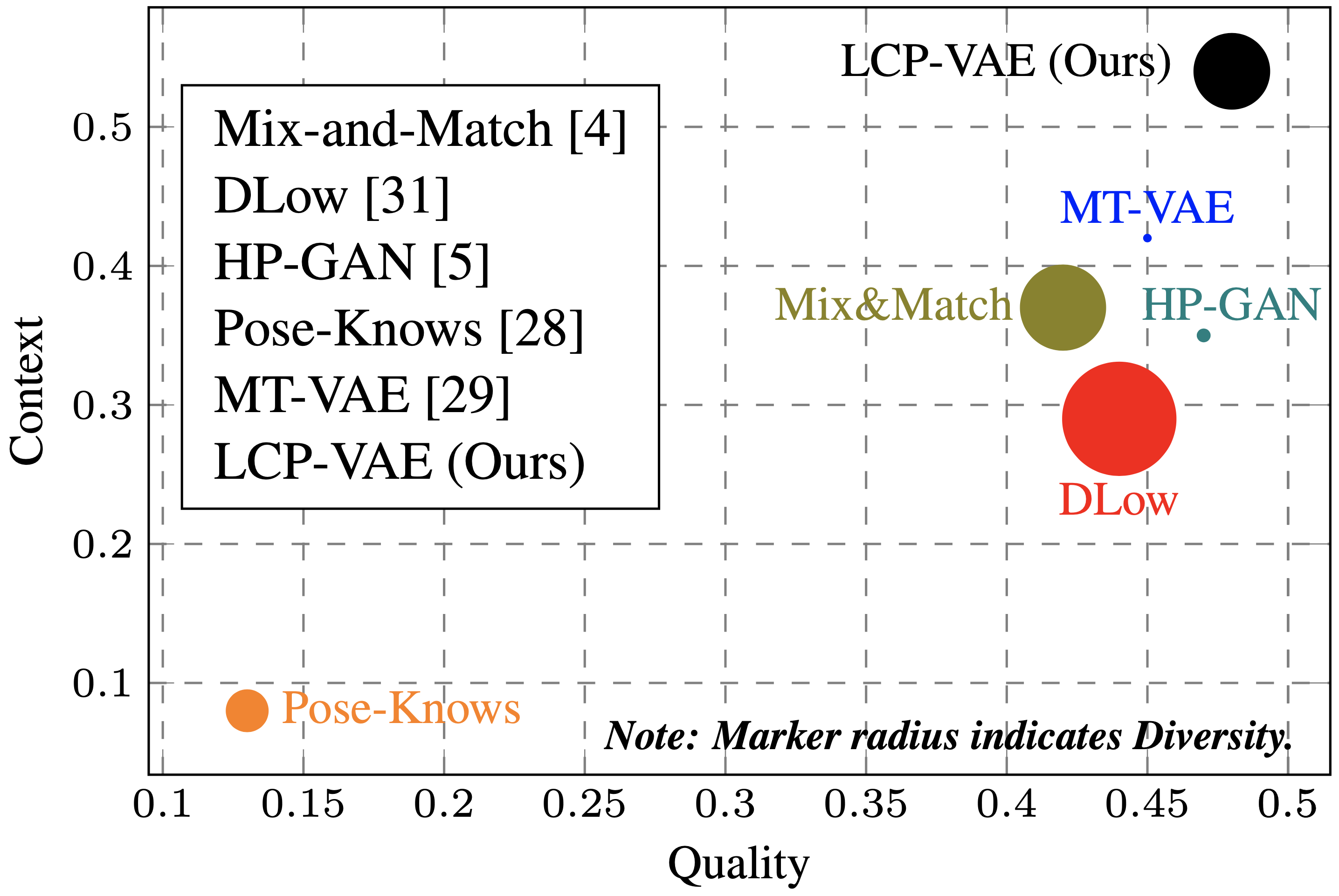}
    \caption{Evaluating the quality, diversity, and context of 
    stochastic motion prediction models. We show each model as a circle in a Context vs. Quality plot, further indicating Diversity with the radius of the marker. These three metrics are defined in Section~\ref{sec:experiments}. 
    Our LCP-VAE approach (in black) yields diverse predictions of higher-quality  than those of other methods, while better preserving the context of the observations.
    }
    \vspace{-10pt}
    \label{fig:qcd}
\end{figure}

Recent attempts to account for motion stochasticity mostly rely on combining a random vector with an encoding of the observed pose sequence~\cite{aliakbarian2019MixAndMatch,barsoum2018hp,butepage2018anticipating,kundu2018bihmp,lin2018human,walker2017pose,yan2018mt,yuan2020dlow}. In particular, the state-of-the-art approaches to diverse human motion prediction~\cite{aliakbarian2019MixAndMatch,yan2018mt,yuan2020dlow} use  conditional variational autoencoders (CVAEs). Here, we argue that standard CVAEs are ill-suited to generate motions that are diverse \textit{and} contextually plausible for the following reasons: 
First, standard CVAEs struggle at capturing diversity
when the conditioning signal 
is highly informative.
In the particular case of human motion prediction, the observed 3D motion (\textit{i.e.}, the condition) contains sufficient signal for 
an expressive motion decoder to generate a natural continuation given only this condition~\cite{aliakbarian2019MixAndMatch,yan2018mt}. 
A typical conditioning scheme, such as concatenating a random latent vector to the condition, then allows the model to learn to ignore the latent variable and only focus on the condition to minimize the reconstruction loss. 
Second, while ignoring the latent variable can be prevented by replacing the traditional deterministic conditioning scheme with a stochastic one~\cite{aliakbarian2019MixAndMatch}, 
capturing context in the diverse predictions with CVAEs is impeded by their use of a general prior on the latent variable. Since this prior is independent of the conditioning signal, during inference, nothing encourages the latent variable to be drawn from a region of the latent space corresponding to the observed motion. In other words, the latent variable is sampled independently of the condition and thus may not carry contextual information about the observed motion.

In this paper, we address these weaknesses by explicitly conditioning the sampling of the latent variable on the past observation, thus encouraging the latent variable to encode relevant information. We will show that this further allows us to depart from the traditional deterministic conditioning scheme, and thus facilitate generating both diverse \emph{and} contextually plausible motions. 
As demonstrated in Fig.~\ref{fig:qcd}, our experiments show that our approach not only yields much higher quality of diverse motions compared to the state-of-the-art stochastic motion prediction methods, but also better preserves the contextual and semantic information of the condition, such as the type of action performed by the person, without explicitly exploiting this information. 

\section{Related Work}
Most motion prediction methods are based on \emph{deterministic} models~\cite{fragkiadaki2015recurrent,ghosh2017learning,gui2018adversarial,gui2018few,jain2016structural,martinez2017human,pavllo2019modeling,pavllo2018quaternet,wei2019motion}, casting motion prediction as a regression task where only one outcome is possible given the observation. While this may produce accurate predictions, it fails to reflect the stochastic nature of human motion, where multiple futures can be highly likely for a single series of past observations. Modeling stochasticity is the topic of this paper, and we therefore focus the discussion on
%the other
the methods that have attempted to do so.

The general trend to incorporate variations in the predicted motions consists of combining information about the observed pose sequence with a random vector. In this context, two types of approaches have been studied: The techniques that directly incorporate the random vector into the motion decoder (typically an RNN) and those that make use of an additional CVAE. 
In the first class of methods,~\cite{lin2018human} samples a random vector $z_t\sim\mathcal{N}(0,I)$ at each time step and concatenates it to the pose input to the RNN decoder. By relying on different and independent random vectors at each time step, however, this strategy is prone to generating discontinuous motions. 
To overcome this,~\cite{kundu2018bihmp} makes use of a single random vector to generate the entire sequence. 
As we will show in our experiments, by relying on concatenation, these two methods contain parameters that are specific to the random vector, and thus give the model the flexibility to ignore this information.
%~\cite{aliakbarian2019MixAndMatch}. \MS{I don't see why we cite your previous work here.}
In~\cite{barsoum2018hp}, instead of using concatenation, the random vector is added to the hidden state produced by the RNN encoder. While addition prevents having parameters that are specific to the random vector, this vector is first transformed by multiplication with a learnable parameter matrix, and thus can again be zeroed out so as to remove the source of diversity, as observed in our experiments. 
\begin{figure*}
    \centering
    \begin{tabular}{c@{ }c@{ }c}
        \includegraphics[width=.31\textwidth]{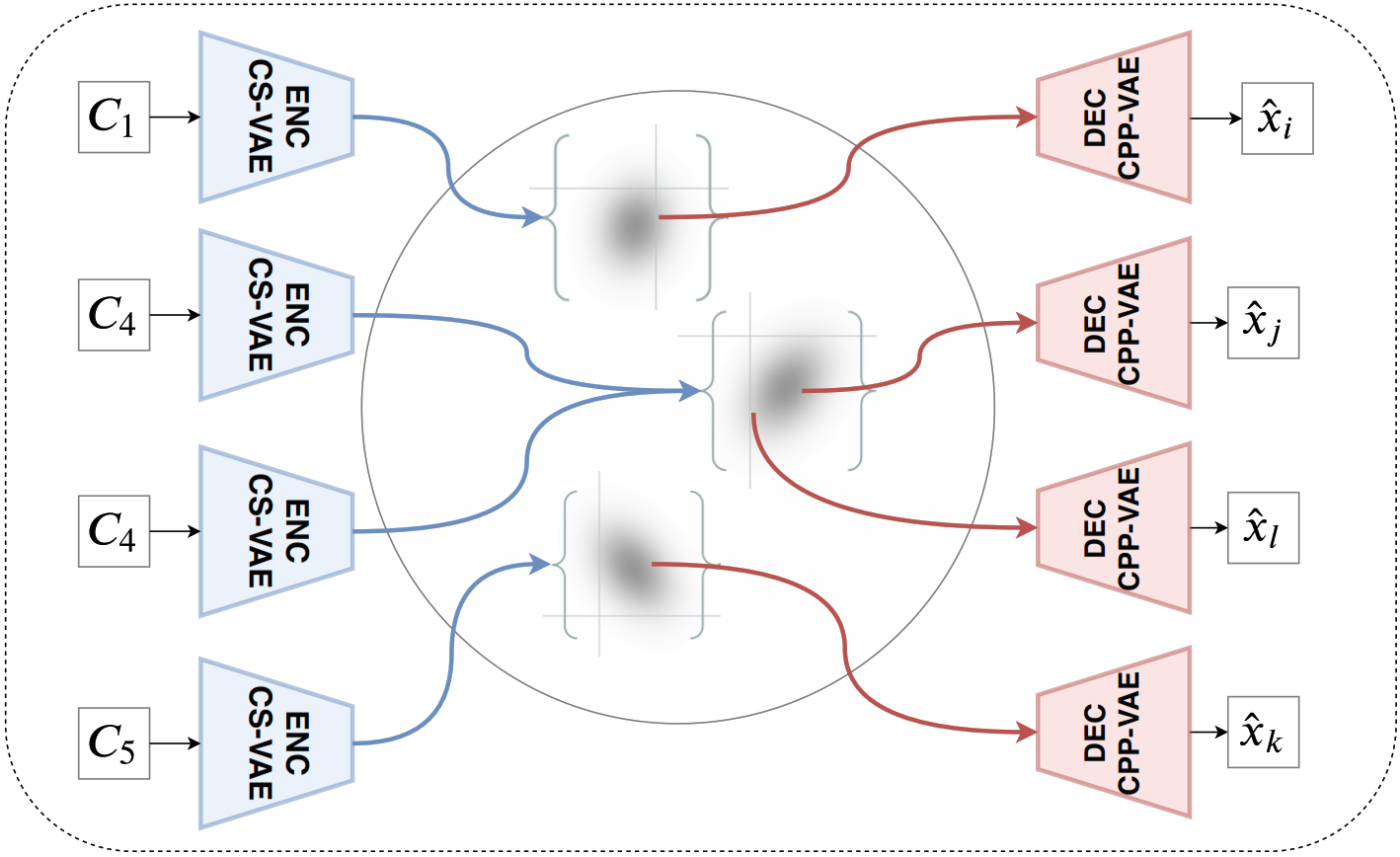} & 
        \includegraphics[width=.33\textwidth]{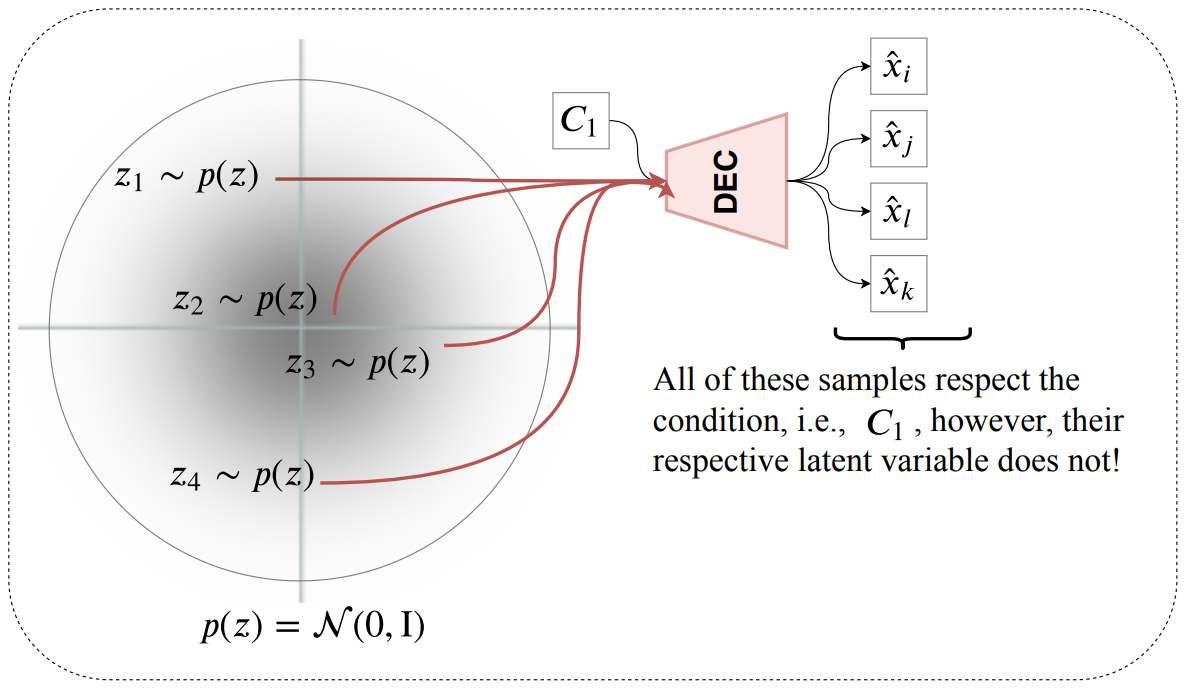} & 
        \includegraphics[width=.31\textwidth]{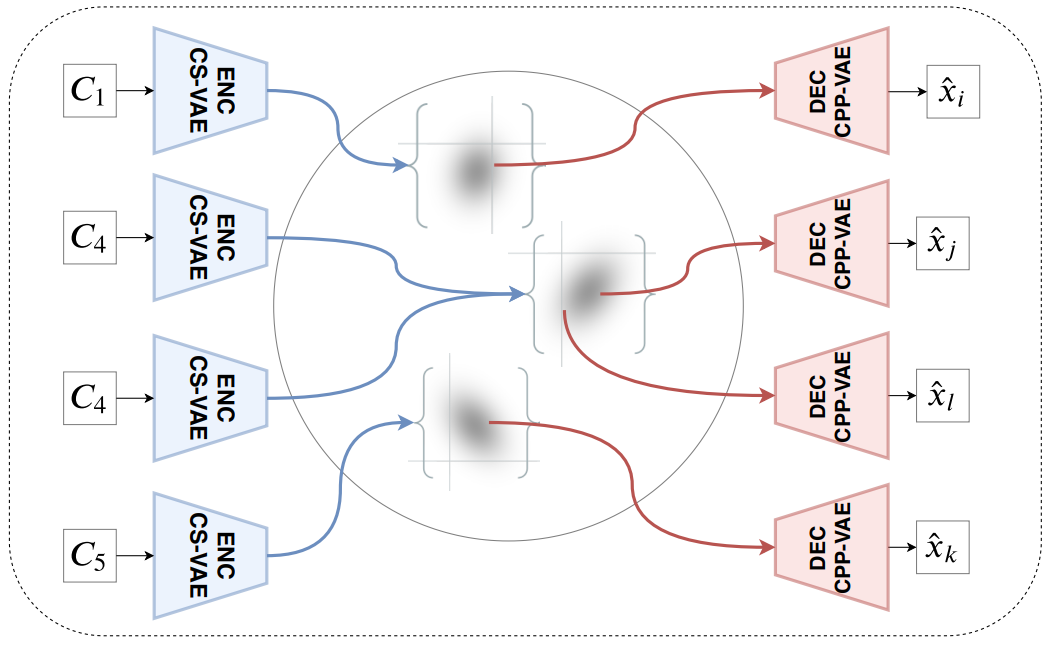} \\
        (a) CVAE at training & (b) CVAE at inference & (c) \texttt{LCP-VAE} (ours) at inference
    \end{tabular}
    % \vspace{5pt}
    \caption{Training and inference for a standard CVAE versus inference for our of {\tt LCP-VAE}.
    (a) During the training of a standard CVAE, the encoder takes as input the combination of the data and the corresponding condition and compresses it into the latent space. The decoder then samples a latent variable from the approximate posterior computed by the encoder, combines the latent variable with  the  conditioning  signal,  and  reconstructs  the  data.  
    (b) During inference in a standard CVAE, the decoder samples different latent  variables  from  the  prior  distribution,  combines  them  with  the  condition, and generates samples that all respect the conditioning signal. 
    However, there is no guarantee that the latent variable is sampled from the region of the prior that corresponds to the given condition.
    (c) In \texttt{LCP-VAE}, at inference time, we utilize an additional encoder acting on the conditioning signal to approximate the posterior of each condition. To generate a sample given a condition, our decoder then samples a latent variable from the posterior of its condition, instead of using a general prior distribution as in CVAEs, and generates a sample. 
    }
    \label{fig:condition_inference}
\end{figure*}

The second category of stochastic methods introduce an additional CVAE between the RNN encoder and decoder. 
In this context,~\cite{walker2017pose} proposes to directly use the pose as conditioning variable. As will be shown in our experiments, while this approach is able to maintain some degree of diversity, albeit less than ours, it yields motions of lower quality because of its use of independent random vectors at each time step. 
Instead of perturbing the pose,~\cite{yan2018mt} uses the RNN encoder's hidden state as conditioning variable in the CVAE, concatenating it with the latent variable. While this approach generates high-quality motions, it suffers from the fact that the CVAE decoder gives the model the flexibility to ignore the random vector, which therefore yields low-diversity outputs. 
To overcome this,~\cite{aliakbarian2019MixAndMatch} perturbs the hidden states via a stochastic Mix-and-Match operation instead of concatenation. Such a perturbation prevents the decoder from decoupling the noise and the condition. However, since the perturbation is not learned and is a non-parametric operation, the quality and the context of the generated motion are inferior to those obtained with our approach.
Similarly to Mix-and-Match~\cite{aliakbarian2019MixAndMatch}, DLow~\cite{yuan2020dlow} aims to modify the sampling process of a fixed, pretrained CVAE so as to ensure diversity in the output space. While this approach is successful at generating diverse motions, it relies on a two-stage training process, making it impossible to \textit{learn} diversity in motions in an end-to-end manner. Furthermore, diversity is achieved by learning a fixed number of transformations, thus preventing this method to generalize to arbitrary numbers of motion modes. Finally,~\cite{yuan2020dlow} does not account for the fact that the diverse continuations of an observation should preserve its context and semantics.
More importantly, all of the above-mentioned CVAE-based approaches use priors that are independent of the condition. We will show in our experiments that such designs are ill-suited for human motion prediction. By contrast, our approach \textit{learns} a conditional prior and is thus able to generate diverse motions of higher quality, carrying the contextual information of the conditioning signal.

\section{Motivation}
\label{sec:motivation}
In essence, VAEs utilize neural networks to learn the distribution of the data. To this end, VAEs first learn to generate a latent variable $z$ given the data $x$, \textit{i.e.}, approximate the posterior distribution $q_\phi(z|x)$, where $\phi$ are the parameters of a neural network, the encoder, whose goal is to model the variation of the data. From this latent random variable $z$, VAEs then generate a new sample $x$ by learning $p_\theta(x|z)$, where $\theta$ denotes the parameters of another neural network, the decoder, whose goal is to maximize the log likelihood of the data. These two networks, \textit{i.e.}, the encoder and the decoder, are trained jointly, using a prior over the latent variable. By using a variational approximation of the posterior, training translates to maximizing the variational lower bound of the log likelihood with respect to the parameters $\phi$ and $\theta$, given by
\begin{align}
    \log p_\theta(x) \geq \mathbb{E}_{q_\phi(z|x)}\Big[\log p_\theta(x|z)\Big] - KL\Big(q_\phi(z|x) \,||\, p(z)\Big)\;,
\end{align}
where the second term on the right hand side encodes the KL divergence between the posterior $q_\phi(z|x)$ and a chosen prior distribution $p(z)$. As an extension to VAEs, CVAEs use auxiliary information, \textit{i.e.}, the conditioning variable or observation, to generate the data $x$. In the standard setting, both the encoder and the decoder are conditioned on the conditioning variable $c$. That is, the encoder becomes $q_\phi(z|x,c)$ and the decoder $p_\theta(x|z,c)$. Then, in theory, the objective of the model should become 
\begin{align}
    \log p_\theta(x|c) \geq \mathbb{E}_{q_\phi(z|x,c)}\Big[\log p_\theta(x|z,c)\Big] \nonumber \\ - KL\Big(q_\phi(z|x,c) \,||\, p(z|c)\Big)\;.
    \label{eq:cvae_elbo}
\end{align}

In practice, however, \textit{the prior distribution of the latent variable is still assumed to be independent of $c$, \textit{i.e.}, $p(z \mid c) = p(z)$.} As illustrated by Fig.~\ref{fig:condition_inference}, at test time, this translates to sampling a latent variable from a region of the prior that may not be correlated with 
the observed condition.

In this paper, we overcome this limitation by explicitly making the sampling of the latent variable depend on the condition. In other words, instead of using $p(z)$ as prior distribution, we truly use $p(z|c)$. This not only respects the theory behind the design of CVAEs, but, as we empirically demonstrate, leads to generating motions of higher quality, that preserve the context of the conditioning signal, \textit{i.e.}, the observed past motion.  To achieve this, we develop a CVAE architecture that learns a distribution not only of the latent variable but also of the conditioning one. We then use this distribution as a prior over the latent variable, making its sampling explicitly dependent on the condition.

\begin{figure*}[t]
\scriptsize
\centering
\begin{tabular}{cc}
\includegraphics[width=0.7\textwidth]{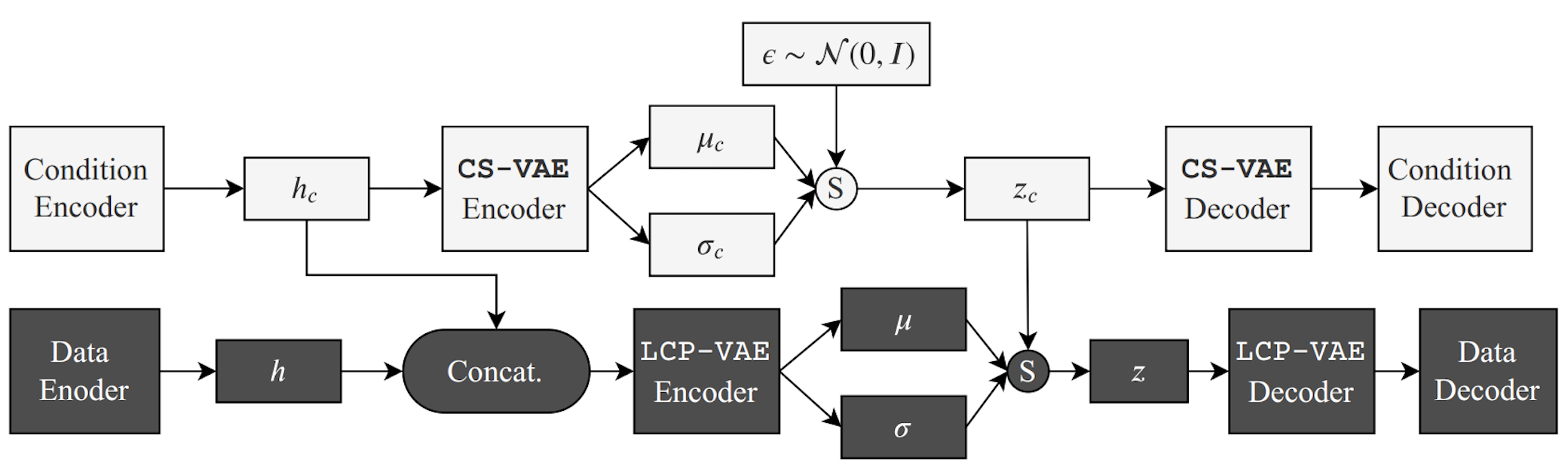} &
\includegraphics[width=0.18\textwidth]{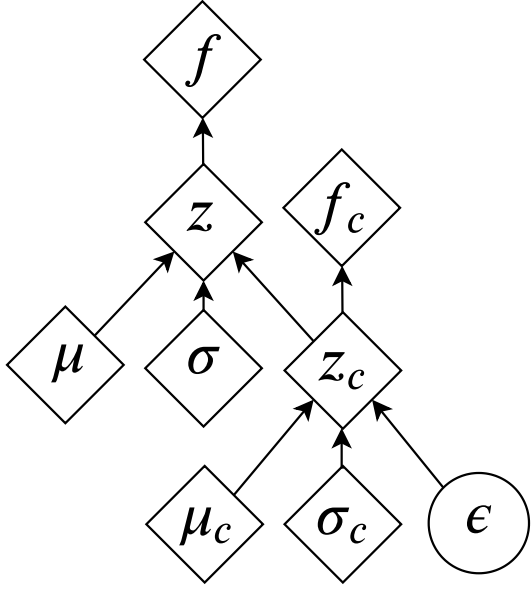}\\
(a) The \texttt{LCP-VAE} framework & (b) Extended Reparameterization Trick
\end{tabular}
\caption{
In an \texttt{LCP-VAE} (a), the sampling of $z$ is conditioned on the CVAE condition via the latent variable $z_c$. Specifically, the posterior distribution of the condition acts as prior on the data posterior. This corresponds to the \textit{extended reparameterization trick} (b) illustrated on the right.
Note that this approximate posterior is not normally distributed anymore.}
\label{fig:method}
\end{figure*}{}

\section{Our Approach: LCP-VAE}
\label{sec:approach}

In this section, we describe our approach to generating diverse and plausible continuations of 3D human pose sequences, where the latent variables are sampled from an appropriate region of the prior distribution. We dub it {\tt LCP-VAE} for VAEs with {\tt L}earned {\tt C}onditional {\tt P}riors. In essence, our framework consists of two autoencoders, one acting on the past observed sequence of 3D poses (\textit{i.e.}, the conditioning signal) and the other on the future sequence of 3D poses (\textit{i.e.}, the continuation of the motion, also referred to as the data) which we wish to model. The latent representation of the condition then serves as a conditioning variable to generate motions from a learned distribution.

As discussed in Section~\ref{sec:motivation}, our approach forces the sampling of the random latent variable to depend on the conditioning one. By making this dependency explicit, we (1) sample an informative latent variable \textit{given} the condition, and thus generate motions of higher quality that preserve the context of the observed motion, and (2) prevent the network from ignoring the latent variable in the presence of a strong condition, thus enabling it to generate diverse outputs.

Note that conditioning the VAE \emph{encoder} via standard strategies, \textit{e.g.}, concatenation, is perfectly fine, since the two inputs to the encoder, \textit{i.e.}, the data and the condition, are deterministic and useful to compress the sample into the latent space. However, conditioning the VAE \emph{decoder} requires special care, which is what we focus on below.

\subsection{Stochastically Conditioning the Decoder} 

We propose to make the sampling of the latent variable 
explicitly depend on the condition instead of treating these two variables as independent. To this end, we first learn the distribution of the condition via a simple VAE, which we refer to as \texttt{CS-VAE} because it acts on the conditioning signal. The goal of \texttt{CS-VAE} is to reconstruct the condition, \textit{i.e.}, the observed past motion, given its latent representation. We take the prior of \texttt{CS-VAE} to be a standard Normal distribution $\mathcal{N}(0,I)$. Following~\cite{kingma2013auto}, this allows us to approximate the \texttt{CS-VAE} posterior with another 
Normal distribution 
via the reparametrization trick expressed as
\begin{align}
    z_c = \mu_c + \sigma_c \odot \epsilon\;,
    \label{eq:observation_reparam}
\end{align}
where $\epsilon\sim \mathcal{N}(0,I)$, and $\mu_c$ and $\sigma_c$ are the parameter vectors of the posterior distribution generated by the VAE encoder, and thus  $z_c\sim\mathcal{N}(\mu_c, \text{diag}(\sigma_c)^2)$.

Following the same strategy for the data VAE, that acts on the future motion,
translates to 
assuming independence of the conditioning and the data,
which we seek to avoid. Therefore, as illustrated in Fig.~\ref{fig:method} (Bottom), we instead define the \texttt{LCP-VAE} posterior not as directly normally distributed but conditioned on the posterior of \texttt{CS-VAE}. To this end, we extend the standard reparameterization trick as 
\begin{align}
    z =  & \mu + \sigma \odot z_c  \nonumber \\ = & \mu + \sigma \odot (\mu_c + \sigma_c \odot \epsilon) \nonumber \\ = & \underbrace{(\mu + \sigma\odot\mu_c)}_\text{\texttt{LCP-VAE}'s mean} \,+\!\! \underbrace{(\sigma\odot\sigma_c)}_\text{\texttt{LCP-VAE}'s std.} \!\!\odot \;\epsilon\,,
    \label{eq:future_reparam}
\end{align}
where $z_c$ comes from Eq.~\ref{eq:observation_reparam}, and $\mu$ and $\sigma$ are the parameter vectors generated by the \texttt{LCP-VAE} encoder. In fact, $z_c$ in Eq.~\ref{eq:observation_reparam} is a sample from the scaled and translated version of $\mathcal{N}(0,I)$ given $\mu_c$ and $\sigma_c$, and $z$ in Eq.~\ref{eq:future_reparam} is a sample from the scaled and translated version of $\mathcal{N}(\mu_c,\text{diag}(\sigma_c)^2)$ given $\mu$ and $\sigma$. Since we have access to the observations during both training and testing, we always sample $z_c$ from the condition posterior. As $z$ is sampled given $z_c$, one expects the latent variable $z$ to carry information about the strong condition, and thus a sample generated from $z$ to correspond to a plausible sample given the condition. This extended reparameterization trick lets us sample one single informative latent variable that contains information about both the data and the conditioning signal. This further allows us to avoid conditioning the \texttt{LCP-VAE} decoder by concatenating the latent variable with a deterministic representation of the condition.
Note that our sampling strategy changes the variational family of the \texttt{LCP-VAE} posterior. In fact, the posterior is no longer $\mathcal{N}(\mu,\text{diag}(\sigma)^2)$, but a Gaussian distribution with mean  $\mu + \sigma\odot\mu_c$ and  covariance matrix $\text{diag}(\sigma\odot\sigma_c)^2$. This will be accounted for when designing the KL divergence loss discussed below.

\subsection{Learning}
To learn the parameters of our model, we rely on the availability of a dataset $D=\{X_1, X_2, ..., X_N\}$ containing $N$ training motion sequences $X_i$, $1 \leq i \leq N$. Each motion, $X_i=\{x_i^1, x_i^2, ..., x_i^T\}$, consists of a sequence of $T$ poses, and each pose, $x_i^t=\{x_{i, 1}^t, x_{i, 2}^t, ..., x_{i, J}^t\}$, comprises $J$ joints forming a skeleton. The pose of each joint is represented as a 4D quaternion. Each training sample $X_i$ contains a past observed motion, or condition, $x_i^{1:t}$, and a future motion, $x_i^{t+1:T}$. For \texttt{CS-VAE}, which learns the distribution of the condition, we define the loss as the KL divergence between its posterior and the standard Gaussian prior, \textit{i.e.},
\begin{align}
    \mathcal{L}_{prior}^{\texttt{CS-VAE}} & = KL\Big(\mathcal{N}(\mu_c, \text{diag}(\sigma_c)^2) \Big\| \mathcal{N}(0,I)\Big) \nonumber \\ & =  -\frac{1}{2}\sum_{j=1}^d \Big(1+\log(\sigma_{c_j}^2) - \mu_{c_j}^2 - \sigma_{c_j}^2 \Big)\;,
    \label{eq:cs-kl-loss}
\end{align}
where $d$ is the dimension of the latent variable $z_c$. By contrast, for \texttt{LCP-VAE}, we define the loss as the KL divergence between the posterior of \texttt{LCP-VAE} and the posterior of  \texttt{CS-VAE}, \textit{i.e.}, of the condition. To this end, at each training iteration, we freeze the weights of \texttt{CS-VAE} before computing the KL divergence, since we do not want to move the posterior of the condition but that of the data. The KL divergence is then computed as the divergence between two multivariate Normal distributions, which yields
\begin{align}
    &\mathcal{L}_{prior}^{\texttt{LCP-VAE}} = \nonumber \\ & KL\Big(\mathcal{N}(\mu + \sigma\odot\mu_c, \text{diag}(\sigma\odot\sigma_c)^2) \Big\| \mathcal{N}(\mu_c, \text{diag}(\sigma_c)^2)\Big)\;.
    \label{eq:cpp-kl-loss}
\end{align}
Let $\Sigma=\text{diag}(\sigma)^2$,  $\Sigma_c=\text{diag}(\sigma_c)^2$, $d$ be the dimensionality of the latent space and $tr\{\cdot\}$ the trace of a square matrix. The loss in Eq.~\ref{eq:cpp-kl-loss} can then be written as\footnote{See the supplementary material for more detail on the KL divergence between two multivariate Gaussians and the derivation of Eq.~\ref{eq:cpp-vae-simplified-kl}.}
\begin{align}
    \mathcal{L}_{prior}^{\texttt{LCP-VAE}} & = -\frac{1}{2}\Big[\log\frac{1}{|\Sigma|}-d + 
    tr\{\Sigma\}  \nonumber \\ & + (\mu_c-(\mu+\Sigma\mu_c))^T\Sigma_c^{-1}(\mu_c-(\mu+\Sigma\mu_c))\Big]\;. 
    \label{eq:cpp-vae-simplified-kl}
\end{align}

After computing the loss in Eq.~\ref{eq:cpp-vae-simplified-kl}, we unfreeze \texttt{CS-VAE} and update it with 
% its previous
the gradient of the loss in Eq.~\ref{eq:cs-kl-loss}.
Trying to match the posterior of \texttt{LCP-VAE} to that of \texttt{CS-VAE} allows us to effectively use our extended reparameterization trick in Eq.~\ref{eq:future_reparam}. Furthermore, we use the standard reconstruction loss for both \texttt{CS-VAE} and \texttt{LCP-VAE}, thus minimizing the mean squared error (MSE) 
\begin{align}
    \mathcal{L}_{rec}^{.} = -\sum_{k=t'}^{T'}\sum_{j=1}^{J}{\|\hat{x}^{k}_{i,j} - x^{k}_{i,j}\|}^2\;,
\label{eq:loss_quat1}
\end{align}
in which for $\mathcal{L}_{rec}^{\texttt{CS-VAE}}$, $t'=1$ and $T'=t$ (\textit{i.e.}, the observation), and for $\mathcal{L}_{rec}^{\texttt{LCP-VAE}}$, $t'=t+1$ and $T'=T$ (\textit{i.e.}, the future motion).
Thus, our complete loss is 
\begin{align}
\mathcal{L} =  \lambda(\mathcal{L}_{prior}^{\texttt{CS-VAE}} + \mathcal{L}_{prior}^{\texttt{LCP-VAE}}) + \mathcal{L}_{rec}^{\texttt{CS-VAE}} + \mathcal{L}_{rec}^{\texttt{LCP-VAE}} \;.
\label{eq:stochastic_loss}
\end{align}
In practice, since the nature of our data is sequential, 
we use a recurrent model for the \texttt{LCP-VAE} encoder. Thus, following the standard practice in diverse human motion prediction~\cite{aliakbarian2019MixAndMatch,walker2017pose,yan2018mt}, we weigh the KL divergence terms by a function $\lambda$ corresponding to the KL annealing weight of~\cite{bowman2015generating}. We start from $\lambda=0$, forcing the model to encode as much information in $z$ as possible, and gradually increase it to $\lambda=1$ during training, following a logistic curve. We then continue training with $\lambda=1$.

In short, our method can be interpreted as a simple yet effective CVAE framework 
for altering the variational family of the posterior such that (1) a latent variable from this posterior distribution is explicitly sampled given the condition, both during training and inference (as illustrated in Fig.~\ref{fig:condition_inference}), and (2) the model is much less likely to suffer from posterior collapse because the mismatch between the posterior and prior distributions makes it harder for learning to drive the KL divergence of Eq.~\ref{eq:cpp-vae-simplified-kl} towards zero.

\section{Experiments}
\label{sec:experiments}
In this section, we compare our approach to the state of the art and evaluate the influence of the different components of our method. 
We provide the implementation details of our approach in the supplementary material. 

\paragraph{Datasets.}
For our experiments, we use the Human3.6M~\cite{h36m_pami}, CMU MoCap\footnote{Available at \texttt{http://mocap.cs.cmu.edu/}.} and Penn Action~\cite{zhang2013actemes} datasets.
Human3.6M comprises more than 800 long indoor motion sequences performed by 11 subjects, leading to 3.6M frames. Each frame contains a person annotated with 3D joint positions and rotation matrices for all 32 joints. In our experiments, for our approach and the replicated VAE-based baselines, we represent each joint in 4D quaternion space. We follow the standard preprocessing and evaluation settings used in~\cite{aliakbarian2019MixAndMatch,gui2018adversarial,jain2016structural,martinez2017human,pavllo2018quaternet}.
The CMU MoCap dataset is another large-scale motion capture dataset covering diverse human activities, such as jumping, running, walking, and playing basketball. Each frame contains a person annotated with 3D joint rotation matrices for all 38 joints. As for Human3.6M, and following standard practice~\cite{li2018convolutional,wei2019motion}, we represent each joint in 4D quaternion space.
The real-world Penn Action dataset~\cite{zhang2013actemes} contains 2326 sequences of 15 different actions, where for each person, 13 joints are annotated in 2D space. For this dataset, we therefore tackle the task of 2D motion prediction. The results on Penn Action are provided in the supplementary material.

\begin{table}[t]
    \centering
    % \small
    \caption{Comparison of \texttt{LCP-VAE} with the state-of-the-art stochastic motion prediction methods. }
    \small
    \scalebox{0.95}{
    \begin{tabular}{l @{ }@{ } c@{ }@{ } c@{ }@{ } c@{ }@{ } c}
    
    \toprule
    \multicolumn{5}{c}{\textbf{Results on Human3.6M}}\\
    \midrule
     & Test MSE  (KL) & Diversity & Quality & Context\\
     Method & \scriptsize{(Reconstructed)} & \scriptsize{(Sampled)} & \scriptsize{(Sampled)} & \scriptsize{(Sampled)} \\
    \midrule
    MT-VAE~\cite{yan2018mt} & 0.51 (0.06) & 0.26 & 0.45 & 0.42\\
    Pose-Knows~\cite{walker2017pose} & 2.08 (N/A) & 1.70 & 0.13 & 0.08\\
    HP-GAN~\cite{barsoum2018hp} & 0.61 (N/A) & 0.48 & 0.47 & 0.35\\
    Mix-and-Match~\cite{aliakbarian2019MixAndMatch} & 0.55 (2.03)  & 3.52 & 0.42 & 0.37\\
    DLow~\cite{yuan2020dlow} & 0.42 (0.61) & \textbf{4.71} & 0.44 & 0.29\\
    \texttt{LCP-VAE} & \textbf{0.41 (8.07)} &  3.12 & \textbf{0.48} & \textbf{0.54}\\
    % \bottomrule
    \\
    \toprule
    \multicolumn{5}{c}{\textbf{Results on CMU MoCap}}\\
    \midrule
     & ELBO  (KL) & Diversity & Quality & Context\\
     Method & \scriptsize{(Reconstructed)} & \scriptsize{(Sampled)} & \scriptsize{(Sampled)} & \scriptsize{(Sampled)}\\
    \midrule
    MT-VAE~\cite{yan2018mt} & 0.25 (0.08) & 0.41 & 0.46 & 0.80 \\
    Pose-Knows~\cite{walker2017pose} & 1.93 (N/A) & 3.00 & 0.18 & 0.27 \\
    HP-GAN~\cite{barsoum2018hp} & 0.24 (N/A) & 0.43 & 0.45 & 0.73\\
    Mix-and-Match~\cite{aliakbarian2019MixAndMatch} & 0.25 (2.92)  & 2.63 & 0.46 & 0.78\\
    DLow~\cite{yuan2020dlow} & 0.26 (0.25) & \textbf{2.90} & 0.44 & 0.34\\
    \texttt{LCP-VAE} & \textbf{0.23 (4.13)} &  2.36 & \textbf{0.48} & \textbf{0.88}\\
    \bottomrule
    \end{tabular}
    }
    \label{tab:stoch}
\end{table}

\begin{table}[t]
\scriptsize
    \caption{Comparison of the generated motions with the ground-truth future motions in terms of context. 
    The gap between the performance of the state-of-the-art pose-based action classifier~\cite{li2018co} with and without true future motions is 0.22 / 0.54 for Human3.6M / CMU MoCap. Using our predictions, this gap decreases to 0.06 / 0.08, showing that our predictions reflect the correct class label.}
    \label{tab:context_upperbound}
    \renewcommand{\arraystretch}{1.2}
    \centering
        \begin{tabular}{l c c c}
    \toprule
    Setting & Obs. & Future Motion & Context (H3.6M / CMU) \\
    \midrule
    Lower bound & GT & Zero velocity & 0.38 / 0.42\\
    Upper bound & GT & GT & 0.60 / 0.96\\
    
    Ours & GT & Sampled from \texttt{LCP-VAE} & 0.54 / 0.88\\
    \bottomrule
    \end{tabular}
\end{table}{}

\begin{figure}
    \centering
    \includegraphics[width=.48\textwidth]{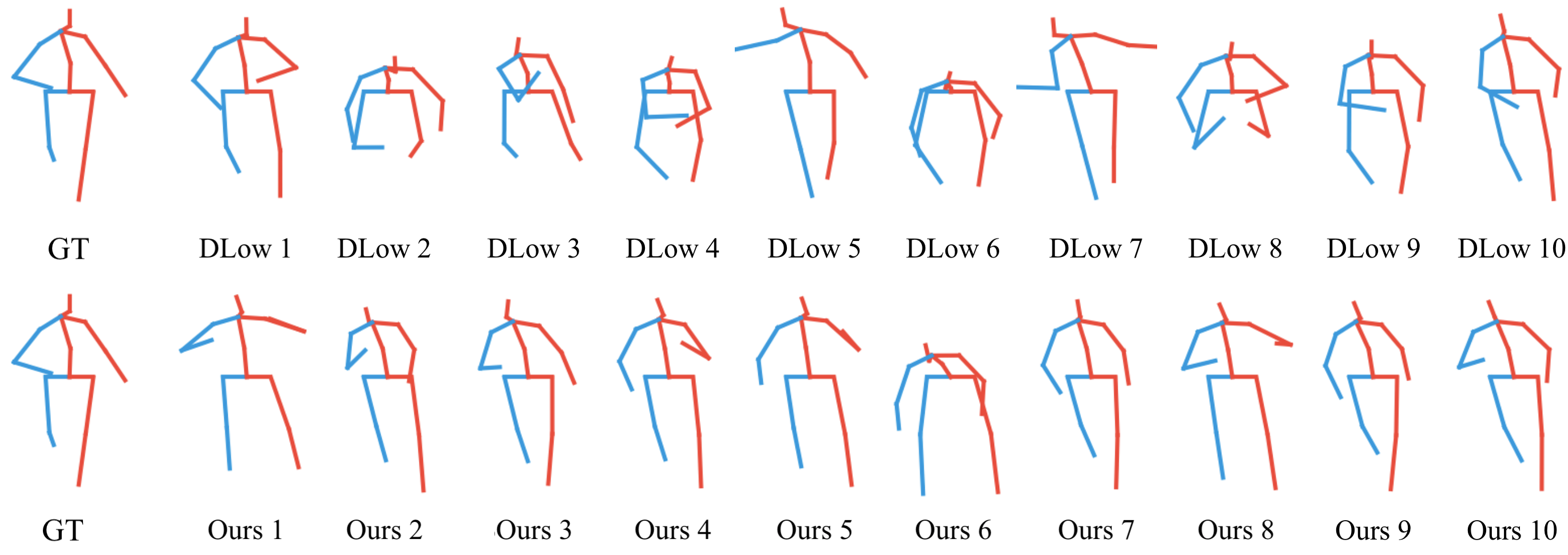}
    \caption{Qualitative comparison of diversity with the state-of-the-art DLow method~\cite{yuan2020dlow}. The first column shows the last time step of the ground-truth motion, and the next 10 columns represent the last time step of 10 different samples from DLow~\cite{yuan2020dlow} (top row) and from our approach (bottom row). While DLow generates highly diverse motions, it loses the context of the observed motion (\textit{walking} in this example). By contrast, our approach generates diverse motions that preserve context in most cases, with a few exceptions, such as Sample 6.}
    \label{fig:dlow}
\end{figure}

\paragraph{Evaluation Metrics.}
To quantitatively evaluate our approach and the state-of-the-art stochastic motion prediction methods~\cite{aliakbarian2019MixAndMatch,barsoum2018hp,walker2017pose,yan2018mt,yuan2020dlow}, we report the test reconstruction error, along with the KL-divergence on the held-out test set. Additionally, we report quality~\cite{aliakbarian2019MixAndMatch} and diversity~\cite{aliakbarian2019MixAndMatch,yang2018diversitysensitive,yuan2020dlow} metrics, which should be considered together. 
Specifically, to measure the diversity of the motions generated by a stochastic model, we make use of the average distance between all pairs of the $K$ motions generated from the same observation. To measure quality, following the quality metric of~\cite{aliakbarian2019MixAndMatch}, we train a binary classifier to discriminate real (ground-truth) samples from fake (generated) ones. The accuracy of this classifier on the test set is inversely proportional to the quality of the generated motions. Note that the results of such a classifier were shown in~\cite{aliakbarian2019MixAndMatch} to match those of human evaluation.
Furthermore, we report a context metric measured as the performance of a strong action classifier~\cite{li2018co} trained on ground-truth motions. Specifically, the classifier is tested on each of the $K$ motions generated from each observation. For $N$ observations and $K$ continuations per observation, the accuracy is measured by computing the argmax over each prediction's probability vector, and we report context as the mean class accuracy on the $K\times N$ motions. 
Unless otherwise stated, for all metrics, we use $K=50$ motions per test observation. For all experiments, we follow~\cite{aliakbarian2019MixAndMatch,yan2018mt} and use 16 frames (\textit{i.e.}, 640ms) as observation to generate the next 60 frames (\textit{i.e.}, 2.4sec). 

\begin{figure*}[t]
    \centering
    \scriptsize
    \begin{tabular}{c c}
    \toprule
    \includegraphics[width=0.48\textwidth]{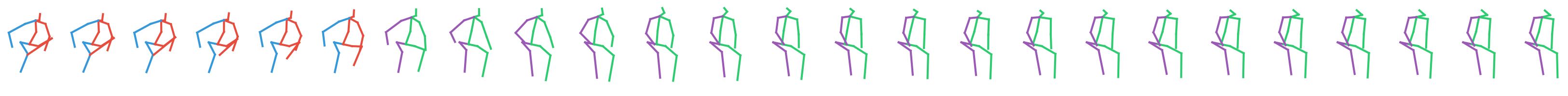} &
    \includegraphics[width=0.48\textwidth]{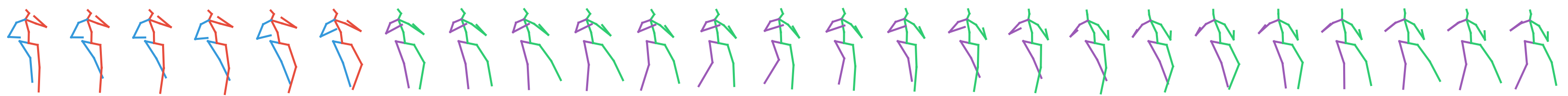} \\
    \midrule
    \includegraphics[width=0.48\textwidth]{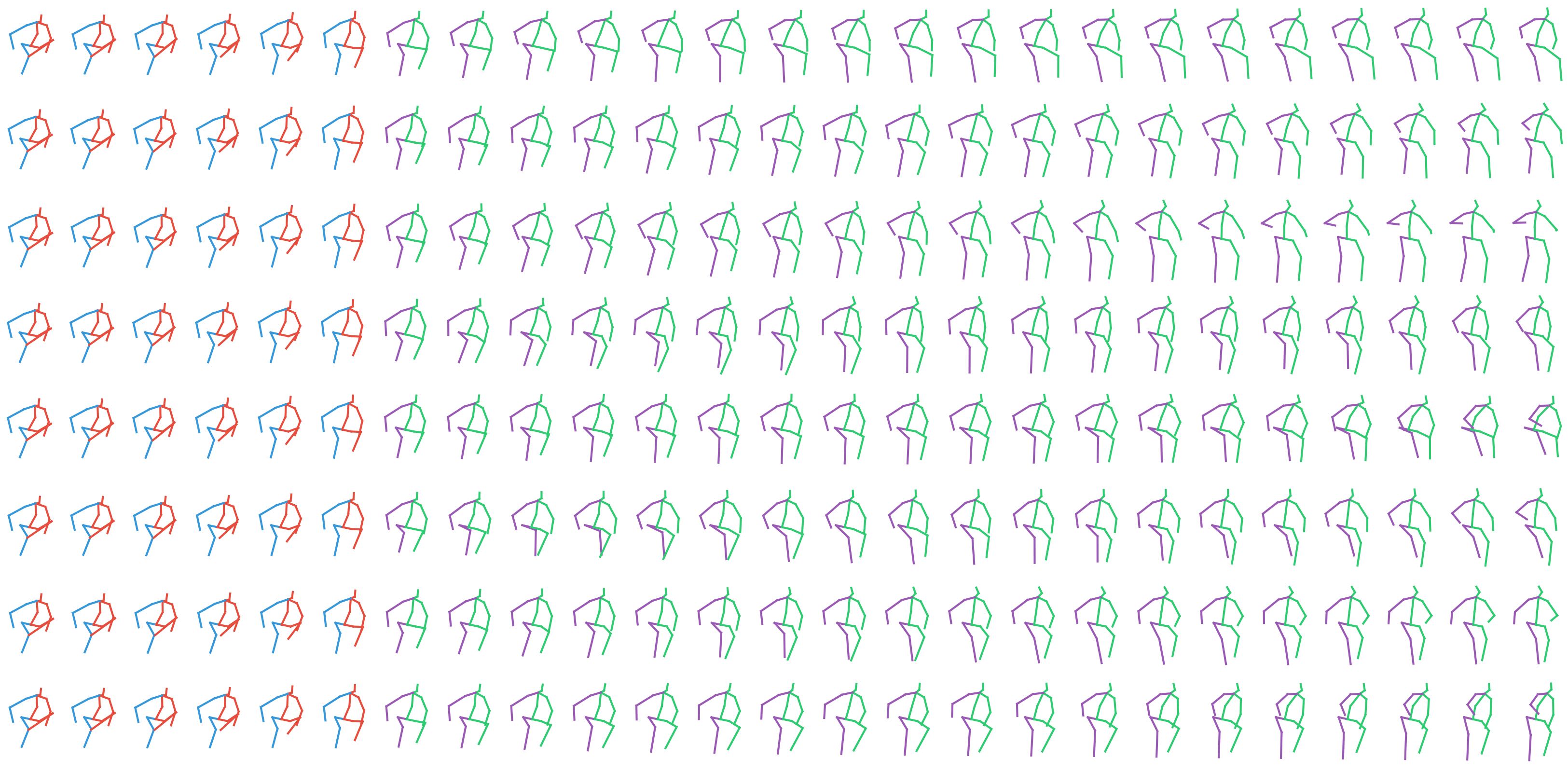} & 
    \includegraphics[width=0.48\textwidth]{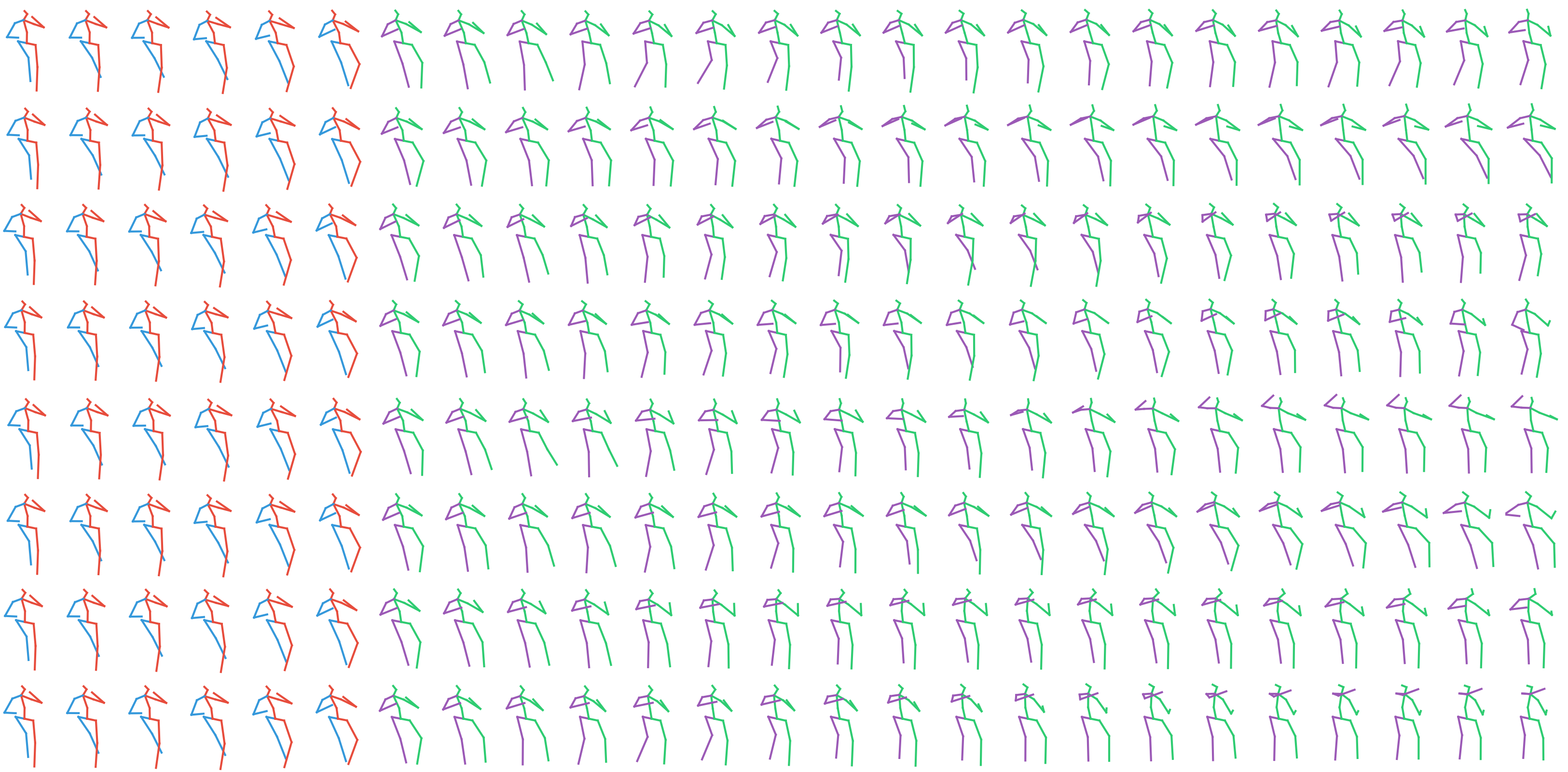} \\
    \bottomrule
    \end{tabular}{}
    \vspace{5pt}
    \caption{Qualitative examples of the diversity in human motion. The first row depicts the ground-truth motion. The first six poses of each row correspond to the observation (the condition) and the remaining ones are sampled from our model. Each row is a randomly sampled motion (not cherry-picked). Note that all motions are natural, with a smooth transition from the observed poses to the generated ones.}
    \label{fig:motion_qualitative}
\end{figure*}{}

\begin{figure*}[t]
    \centering
    \scriptsize
    \begin{tabular}{c@{}c@{}c@{}c}
         \includegraphics[width=0.125\textwidth]{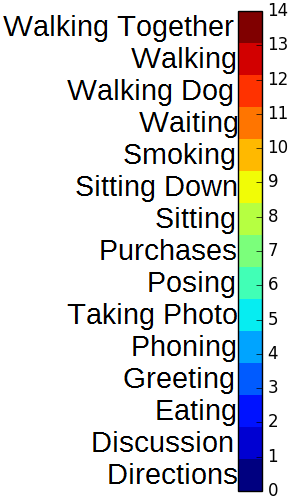} & 
         \includegraphics[width=0.29\textwidth]{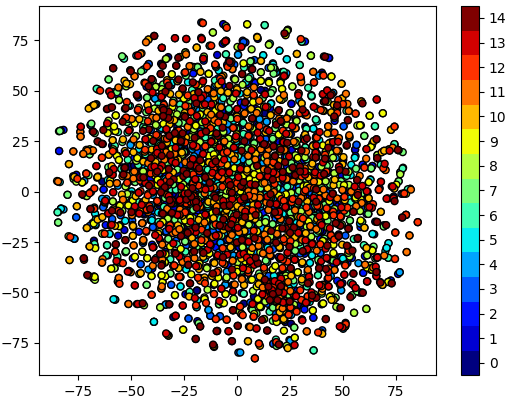} & 
         \includegraphics[width=0.29\textwidth]{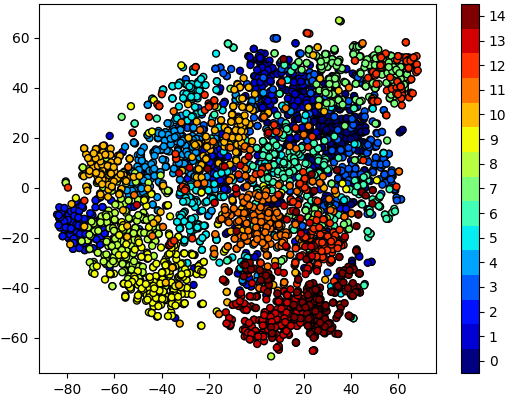} &
         \includegraphics[width=0.29\textwidth]{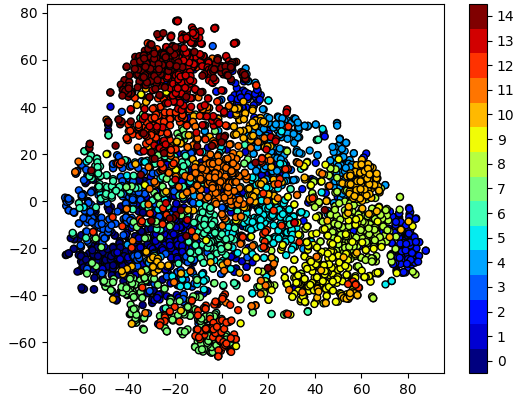} \\
         & MT-VAE~\cite{yan2018mt} & \texttt{LCP-VAE} ($\mu$) & \texttt{LCP-VAE} ($\mu+\mu_c\odot\sigma_c$)\\
    \end{tabular}
    \vspace{5pt}
    \caption{t-SNE plots of the posterior mean for 3750 test motions. With MT-VAE~\cite{yan2018mt}, all classes are mixed, suggesting that the latent variable carries little information about the motion type. By contrast, our condition-dependent sampling allows \texttt{LCP-VAE} to better preserve context. Note that some actions, such as ``Discussion'' and ``Directions'', are very hard to identify and are thus spread over other actions. Others, such as ``Walking'', ``Walking with dog'', and ``Walking together'', or ``Sitting'' and ``Sitting down'', overlap due to their similarity. 
    }
    \label{fig:tsne}
\end{figure*}{}

\subsection{Comparison to the State of the Art} 
In Table~\ref{tab:stoch}, we compare our approach (whose detailed architecture is described in the supplementary material) with the state-of-the-art stochastic motion prediction models~\cite{aliakbarian2019MixAndMatch,barsoum2018hp,walker2017pose,yan2018mt,yuan2020dlow}. Note that, when evaluating stochastic motion prediction models, one should consider the reported metrics jointly to truly evaluate the performance of a stochastic model, as we do in Fig.~\ref{fig:qcd}. For instance, while MT-VAE~\cite{yan2018mt} and HP-GAN~\cite{barsoum2018hp} generate high-quality motions, they are not diverse. Conversely, while Pose-Knows~\cite{walker2017pose} generates diverse motions, they are of low quality. By contrast, our approach generates both high quality and diverse motions. This is also the case of Mix-and-Match~\cite{aliakbarian2019MixAndMatch} and DLow~\cite{yuan2020dlow}, which, however, preserve much less contextual information and semantics of the observed motions. Fig.~\ref{fig:dlow} compares motions generated by our approach and the most diverse model DLow. Note that DLow~\cite{yuan2020dlow} is unable to preserve the contextual information of the observation, whereas our approach does. In fact, none of the baselines effectively conveys the semantics of the observation to the generated motions. As shown in Table~\ref{tab:context_upperbound}, the upper bounds for context on Human3.6M / CMU are 0.60 / 0.96 (\textit{i.e.}, the performance of~\cite{li2018co} given the ground-truth motions), which our method approaches closely, yielding context scores of 0.54 / 0.88 when given only the observed motion. The fact that our method better preserves context is further evidenced by the t-SNE~\cite{maaten2008visualizing} plots of Fig.~\ref{fig:tsne}, where different samples of various actions are better separated for our approach than for MT-VAE~\cite{yan2018mt}, the method that best retains context among the baselines. 
Altogether, as also supported by the qualitative results of Fig.~\ref{fig:motion_qualitative} and in the supplementary material, our approach yields diverse, high-quality and context-preserving predictions.

\begin{table}[t]
\scriptsize
    \centering
        \caption{Comparison with the state-of-the-art stochastic motion prediction models in terms of sampling quality for 4 actions of Human3.6M (all methods use the best of $K=50$ sampled motions).}
    \renewcommand{\arraystretch}{1.3}
\scalebox{0.93}{
    \begin{tabular}{l @{ }@{ } c @{ }@{ } c @{ }@{ } c @{ }@{ } c @{ }@{ } c @{ }@{ } c @{ }@{ } @{ }@{ } c @{ }@{ } c @{ }@{ } c @{ }@{ } c @{ }@{ } c @{ }@{ } c}
\toprule
    & \multicolumn{6}{c}{Walking} & \multicolumn{6}{c}{Eating} \\
    \midrule
    Method & 80  & 160 & 320 & 400 & 560 & 1000  & 80  & 160 & 320 & 400 & 560 & 1000  \\
    \midrule
    
    MT-VAE~\cite{yan2018mt} & 
    0.73 & 0.79 & 0.90 & 0.93 & 0.95 & 1.05 & 
    0.68 & 0.74 & 0.95 & 1.00 & 1.03 & 1.38 \\
    
    HP-GAN~\cite{barsoum2018hp} & 
    0.61 & 0.62 & 0.71 & 0.79 & 0.83 & 1.07 & 
    0.53 & 0.67 & 0.79 & 0.88 & 0.97 & 1.12 \\
    
    Pose-Knows~\cite{walker2017pose} & 
    0.56 & 0.66 & 0.98 & 1.05 & 1.28 & 1.60 & 
    0.44 & 0.60 & 0.71 & 0.84 & 1.05 & 1.54 \\ 
    
    Mix\&Match~\cite{aliakbarian2019MixAndMatch} & 
    0.33 & 0.48 & 0.56 & \underline{0.58} & \underline{0.64} & \textbf{0.68} &
    \underline{0.23} & 0.34 & \underline{0.41} & \textbf{0.50} & \underline{0.61} & \underline{0.91}  \\
    
    DLow~\cite{yuan2020dlow} & \underline{0.31} & \underline{0.42} & \underline{0.53} & 0.75 & 0.83 & 0.96 
    & 0.24 & \underline{0.32} & 0.44 & 0.55 & 0.77 & 0.97 \\
    
    \texttt{LCP-VAE} & 
    \textbf{0.22} & \textbf{0.36} & \textbf{0.47} & \textbf{0.52} & \textbf{0.58} & \underline{0.69} & 
    \textbf{0.19} & \textbf{0.28} & \textbf{0.40} & \underline{0.51} & \textbf{0.58} & \textbf{0.90} \\ 
    \midrule
    & \multicolumn{6}{c}{Smoking} & \multicolumn{6}{c}{Discussion}\\
    \midrule
    Method & 80  & 160 & 320 & 400 & 560 & 1000  & 80  & 160 & 320 & 400 & 560 & 1000  \\
    \midrule
    
    MT-VAE~\cite{yan2018mt} & 
    1.00 & 1.14 & 1.43 & 1.44 & 1.68 & 1.99 & 
    0.80 & 1.01 & 1.22 & 1.35 & 1.56 & 1.69 \\
    
    HP-GAN~\cite{barsoum2018hp} & 
    0.64 & 0.78 & 1.05 & 1.12 & 1.64 & 1.84 & 
    0.79 & 1.00 & 1.12 & 1.29 & 1.43 & 1.71 \\
    
    Pose-Knows~\cite{walker2017pose} & 
    0.59 & 0.83 & 1.25 & 1.36 & 1.67 & 2.03 & 
    0.73 & 1.10 & 1.33 & 1.34 & 1.45 & 1.85 \\
    
    Mix\&Match~\cite{aliakbarian2019MixAndMatch} & 
    \underline{0.23} & \textbf{0.42} & \underline{0.79} & \underline{0.77} & \underline{0.82} & \underline{1.25} & 
    \underline{0.25} & 0.60 & 0.83 & 0.89 & \underline{1.12} & \underline{1.30} \\
    
    DLow~\cite{yuan2020dlow} & \textbf{0.21} & \underline{0.43} & 0.80 & 0.79 & 0.97 & 1.65 & 
    0.31 & \underline{0.55} & \textbf{0.80} & \underline{0.88} & 1.15 & 1.33  \\
    
    \texttt{LCP-VAE} & 
    \underline{0.23} & \underline{0.43} & \textbf{0.77} &\textbf{ 0.75} & \textbf{0.78} & \textbf{1.23} & 
    \textbf{0.21} & \textbf{0.52} & \underline{0.81} & \textbf{0.84} & \textbf{1.04} & \textbf{1.28} \\
    
\bottomrule
    \end{tabular}
    }
    \label{tab:mae_stoch}
\end{table}

\paragraph{Evaluating Sampling Quality.}
To further evaluate the sampling quality, we evaluate stochastic baselines using the standard mean angle error (MAE) metric in Euler space. To this end, we use the best of the $K=50$ generated motions for each observation (referred to as S-MSE in~\cite{yan2018mt}). A model that generates more diverse motions has higher chances of producing a motion  close to the ground-truth one. As shown in Table~\ref{tab:mae_stoch}, this is the case with our approach, Mix-and-Match~\cite{aliakbarian2019MixAndMatch}, and DLow~\cite{yuan2020dlow}\footnote{We used the official implementation of DLow~\cite{yuan2020dlow} (\url{https://github.com/Khrylx/DLow}) and trained it to generate 50 diverse predictions via training 50 latent transformations.}, which all yield high diversity. 
Note, however, that Mix-and-Match~\cite{aliakbarian2019MixAndMatch} and DLow~\cite{yuan2020dlow} focus mainly on encouraging  motion diversity, either by introducing a stochastic conditioning scheme that ignores part of the conditioning signal~\cite{aliakbarian2019MixAndMatch} or via a fixed set of latent transformations that spread the latent variable across different regions of the latent space~\cite{yuan2020dlow}. As such, they inevitably lose some context in the generated motions. By contrast, we  learn a context-preserving latent representation, which helps our approach to generate higher quality motions.
In the supplementary material, we additionally compare our approach with the state-of-the-art deterministic motion prediction models.

\subsection{Ablation Studies and Analysis}
In this section, we analyze the performance of different stochastic motion prediction models. We further provide  insights on different means of conditioning for the encoder and the decoder of our model.

\paragraph{Why existing methods fail to generate diverse \textit{and} plausible motions?}
Here, we further analyze the behavior of different baselines
under different evaluation metrics, reported in Fig.~\ref{fig:qcd} and Table~\ref{tab:stoch}. 
These results show that MT-VAE~\cite{yan2018mt} tends to ignore the random variable $z$, thus ignoring the root of stochasticity. As a consequence, it yields a low diversity, much lower than ours, but produces samples of high quality, albeit almost identical. We empirically traced this back to the magnitude of the weights acting on $z$ being orders of magnitude smaller than that of those acting on the condition, suggesting that MT-VAE~\cite{yan2018mt} has learned to ignore the latent variable.
Similarly, our experiments show that HP-GAN~\cite{barsoum2018hp} also yields limited diversity despite its use of random noise during inference. 
This is due to the fact\footnote{We noticed this by studying and running the authors' publicly available code at \texttt{https://github.com/ebarsoum/hpgan}.} that in~\cite{barsoum2018hp}, $z$ is linearly transformed by a learnable weight matrix $W$ before being added to the motion representation. We then observed that the magnitude of the parameters in $W$ is almost zero, allowing HP-GAN~\cite{barsoum2018hp} to ignore $z$.
Unlike~\cite{barsoum2018hp,yan2018mt}, Pose-Knows~\cite{walker2017pose} produces more diverse motions, but their quality is very low. By inspecting their code\footnote{{\tt https://github.com/puffin444/poseknows}}, we observed that the random vectors that are concatenated to the poses at each time-step are sampled independently of each other, which translates to discontinuities in the generated motions. 
The Mix-and-Match approach~\cite{aliakbarian2019MixAndMatch} yields generated motions with reasonably high diversity and quality. Its architecture is very close to that of MT-VAE~\cite{yan2018mt}, but the deterministic concatenation operation is replaced with a stochastic perturbation of the hidden state with the noise. 
While this prevents the model from ignoring the random noise, the perturbation is not learnt, which translates to lower-quality predictions than ours and that of other baselines. 
DLow~\cite{yuan2020dlow}, which learns to diversify the sampled latent variables of a fixed, pre-trained CVAE through multiple transformations, also generates highly diverse motions. However, the predictions tend to ignore the context of the observed motion as the transformations are trained so as to maximize diversity.
While this indeed makes the DLow predictions diverse, the resulting model relies on a fixed number of learned transformations, and thus cannot generate an arbitrary number of diverse motions at test time. That is, a DLow model with $K$ transformations can generate only $K$ different modes of the data; generating $K+1$ modes would require retraining the model.

\paragraph{Ablation Study on Different Means of Conditioning.}
Finally, we study various designs to condition the VAE encoder and decoder. 
To this end, we employ either a deterministic representation of the observation, $h_t$, or a stochastic one, $z_c$, as conditioning variable. 
We then evaluate the conditioning of the VAE encoder on either of these variables, which, as discussed before, can be done safely via concatenation as both the condition and the data carry useful information to compress the future motion into the latent space. By contrast, for the decoder, we compare the use of $h_t$ or $z_c$ via  either concatenation or  our extended reparameterization trick of Eq.~\ref{eq:future_reparam}.
As evidenced by the first and fourth rows of Table~\ref{tab:arch_design}, the use of a stochastic condition and of our extended reparameterization trick to condition the decoder are both critical to successfully achieve diversity while maintaining context. Furthermore, as shown in the fourth row, for the encoder, a deterministic condition works better than a stochastic one. 
Note that, depending on the application, one may prefer a deterministic condition ($h_t$) over a stochastic one if contextual information in the generated motion matters more than diversity, or a stochastic one ($z_c$) in the opposite situation. In this work, as we aim to generate contextually plausible and diverse motions, we use a deterministic condition for the encoder and a stochastic one for the decoder, which yields a good trade-off between diversity and context, as depicted by the fourth row in Table~\ref{tab:arch_design}.

\begin{table}
    \centering
    \small
    \caption{Evaluation of various architecture designs for a CVAE. 
    }
   \tabcolsep=0.26cm
\scalebox{0.9}{
    \begin{tabular}{c c c c}
    \toprule
        Encoder  & Decoder &  & \\
         Conditioning & Conditioning & Diversity & Context\\
        \midrule
        Concat ($z_c$) & Reparam. ($z_c$) & 3.35 & 0.51\\
        Concat ($h_t$) & Concat ($h_t$) & 0.24 & 0.43\\
        Concat ($z_c$) & Concat ($z_c$)& 1.28 & 0.40\\
        
        Concat ($h_t$) & Reparam ($z_c$) & 3.12 & 0.54\\
        \bottomrule
    \end{tabular}
    }
    \label{tab:arch_design}
    \vspace{-5pt}
\end{table}{}

\section{Conclusion}
In this paper, we have studied the problem of conditionally generating diverse and contextually plausible 3D human motions. We have studied the reasons why existing CVAE-based methods fail to achieve these two goals jointly, and have addressed their weaknesses by forcing the sampling of the latent variable to explicitly depend on the observed motions. 
We have demonstrated that our approach can generate high quality diverse motions that carry out the semantics of the observed motion, such as the type of action performed by the person, without explicitly exploiting this information. 
These properties are highly beneficial to practical applications that can exploit motion predictions, 
such as action anticipation~\cite{aliakbarian2016deep,rodriguez2018action,Aliakbarian_2017_ICCV}, pedestrian intention forecasting~\cite{aliakbarian2018viena2}, 
and human tracking~\cite{alahi2016social,kosaraju2019social,saleh2020artist}.

{\small
\bibliographystyle{ieee_fullname}
\bibliography{cvpr}

\begin{thebibliography}{10}\itemsep=-1pt

\bibitem{alahi2016social}
Alexandre Alahi, Kratarth Goel, Vignesh Ramanathan, Alexandre Robicquet, Li
  Fei-Fei, and Silvio Savarese.
\newblock Social lstm: Human trajectory prediction in crowded spaces.
\newblock In {\em Proceedings of the IEEE conference on computer vision and
  pattern recognition}, pages 961--971, 2016.

\bibitem{aliakbarian2016deep}
Mohammad~Sadegh Aliakbarian, Fatemehsadat Saleh, Basura Fernando, Mathieu
  Salzmann, Lars Petersson, and Lars Andersson.
\newblock Deep action-and context-aware sequence learning for activity
  recognition and anticipation.
\newblock {\em arXiv preprint arXiv:1611.05520}, 2016.

\bibitem{aliakbarian2018viena2}
Mohammad~Sadegh Aliakbarian, Fatemeh~Sadat Saleh, Mathieu Salzmann, Basura
  Fernando, Lars Petersson, and Lars Andersson.
\newblock Viena$^2$: A driving anticipation dataset.
\newblock In {\em Asian Conference on Computer Vision}, pages 449--466.
  Springer, 2018.

\bibitem{aliakbarian2019MixAndMatch}
Sadegh Aliakbarian, Fatemeh~Sadat Saleh, Mathieu Salzmann, Lars Petersson, and
  Stephen Gould.
\newblock A stochastic conditioning scheme for diverse human motion prediction.
\newblock In {\em The IEEE/CVF Conference on Computer Vision and Pattern
  Recognition (CVPR)}, June 2020.

\bibitem{barsoum2018hp}
Emad Barsoum, John Kender, and Zicheng Liu.
\newblock Hp-gan: Probabilistic 3d human motion prediction via gan.
\newblock In {\em Proceedings of the IEEE Conference on Computer Vision and
  Pattern Recognition Workshops}, pages 1418--1427, 2018.

\bibitem{bowman2015generating}
Samuel~R Bowman, Luke Vilnis, Oriol Vinyals, Andrew~M Dai, Rafal Jozefowicz,
  and Samy Bengio.
\newblock Generating sentences from a continuous space.
\newblock {\em arXiv preprint arXiv:1511.06349}, 2015.

\bibitem{butepage2018anticipating}
Judith B{\"u}tepage, Hedvig Kjellstr{\"o}m, and Danica Kragic.
\newblock Anticipating many futures: Online human motion prediction and
  generation for human-robot interaction.
\newblock In {\em 2018 IEEE International Conference on Robotics and Automation
  (ICRA)}, pages 1--9. IEEE, 2018.

\bibitem{chung2014empirical}
Junyoung Chung, Caglar Gulcehre, KyungHyun Cho, and Yoshua Bengio.
\newblock Empirical evaluation of gated recurrent neural networks on sequence
  modeling.
\newblock {\em arXiv preprint arXiv:1412.3555}, 2014.

\bibitem{fragkiadaki2015recurrent}
Katerina Fragkiadaki, Sergey Levine, Panna Felsen, and Jitendra Malik.
\newblock Recurrent network models for human dynamics.
\newblock In {\em Proceedings of the IEEE International Conference on Computer
  Vision}, pages 4346--4354, 2015.

\bibitem{ghosh2017learning}
Partha Ghosh, Jie Song, Emre Aksan, and Otmar Hilliges.
\newblock Learning human motion models for long-term predictions.
\newblock In {\em 2017 International Conference on 3D Vision (3DV)}, pages
  458--466. IEEE, 2017.

\bibitem{gui2018adversarial}
Liang-Yan Gui, Yu-Xiong Wang, Xiaodan Liang, and Jos{\'e}~MF Moura.
\newblock Adversarial geometry-aware human motion prediction.
\newblock In {\em Proceedings of the European Conference on Computer Vision
  (ECCV)}, pages 786--803, 2018.

\bibitem{gui2018few}
Liang-Yan Gui, Yu-Xiong Wang, Deva Ramanan, and Jos{\'e}~MF Moura.
\newblock Few-shot human motion prediction via meta-learning.
\newblock In {\em Proceedings of the European Conference on Computer Vision
  (ECCV)}, pages 432--450, 2018.

\bibitem{h36m_pami}
Catalin Ionescu, Dragos Papava, Vlad Olaru, and Cristian Sminchisescu.
\newblock Human3.6m: Large scale datasets and predictive methods for 3d human
  sensing in natural environments.
\newblock {\em IEEE Transactions on Pattern Analysis and Machine Intelligence},
  36(7):1325--1339, jul 2014.

\bibitem{jain2016structural}
Ashesh Jain, Amir~R Zamir, Silvio Savarese, and Ashutosh Saxena.
\newblock Structural-rnn: Deep learning on spatio-temporal graphs.
\newblock In {\em Proceedings of the IEEE Conference on Computer Vision and
  Pattern Recognition}, pages 5308--5317, 2016.

\bibitem{kingma2013auto}
Diederik~P Kingma and Max Welling.
\newblock Auto-encoding variational bayes.
\newblock {\em arXiv preprint arXiv:1312.6114}, 2013.

\bibitem{kosaraju2019social}
Vineet Kosaraju, Amir Sadeghian, Roberto Mart{\'\i}n-Mart{\'\i}n, Ian Reid,
  Hamid Rezatofighi, and Silvio Savarese.
\newblock Social-bigat: Multimodal trajectory forecasting using bicycle-gan and
  graph attention networks.
\newblock In {\em Advances in Neural Information Processing Systems}, pages
  137--146, 2019.

\bibitem{kundu2018bihmp}
Jogendra~Nath Kundu, Maharshi Gor, and R~Venkatesh Babu.
\newblock Bihmp-gan: Bidirectional 3d human motion prediction gan.
\newblock {\em arXiv preprint arXiv:1812.02591}, 2018.

\bibitem{li2018convolutional}
Chen Li, Zhen Zhang, Wee Sun~Lee, and Gim Hee~Lee.
\newblock Convolutional sequence to sequence model for human dynamics.
\newblock In {\em Proceedings of the IEEE Conference on Computer Vision and
  Pattern Recognition}, pages 5226--5234, 2018.

\bibitem{li2018co}
Chao Li, Qiaoyong Zhong, Di Xie, and Shiliang Pu.
\newblock Co-occurrence feature learning from skeleton data for action
  recognition and detection with hierarchical aggregation.
\newblock {\em arXiv preprint arXiv:1804.06055}, 2018.

\bibitem{lin2018human}
Xiao Lin and Mohamed~R Amer.
\newblock Human motion modeling using dvgans.
\newblock {\em arXiv preprint arXiv:1804.10652}, 2018.

\bibitem{maaten2008visualizing}
Laurens van~der Maaten and Geoffrey Hinton.
\newblock Visualizing data using t-sne.
\newblock {\em Journal of machine learning research}, 9(Nov):2579--2605, 2008.

\bibitem{wei2019motion}
Wei Mao, Miaomiao Liu, Mathieu Salzmann, and Hongdong Li.
\newblock Learning trajectory dependencies for human motion prediction.
\newblock In {\em ICCV}, 2019.

\bibitem{martinez2017human}
Julieta Martinez, Michael~J Black, and Javier Romero.
\newblock On human motion prediction using recurrent neural networks.
\newblock In {\em 2017 IEEE Conference on Computer Vision and Pattern
  Recognition (CVPR)}, pages 4674--4683. IEEE, 2017.

\bibitem{pavllo2019modeling}
Dario Pavllo, Christoph Feichtenhofer, Michael Auli, and David Grangier.
\newblock Modeling human motion with quaternion-based neural networks.
\newblock {\em arXiv preprint arXiv:1901.07677}, 2019.

\bibitem{pavllo2018quaternet}
Dario Pavllo, David Grangier, and Michael Auli.
\newblock Quaternet: A quaternion-based recurrent model for human motion.
\newblock {\em arXiv preprint arXiv:1805.06485}, 2018.

\bibitem{rodriguez2018action}
Cristian Rodriguez, Basura Fernando, and Hongdong Li.
\newblock Action anticipation by predicting future dynamic images.
\newblock In {\em Proceedings of the European Conference on Computer Vision
  (ECCV)}, pages 0--0, 2018.

\bibitem{Aliakbarian_2017_ICCV}
Mohammad Sadegh~Aliakbarian, Fatemeh Sadat~Saleh, Mathieu Salzmann, Basura
  Fernando, Lars Petersson, and Lars Andersson.
\newblock Encouraging lstms to anticipate actions very early.
\newblock In {\em The IEEE International Conference on Computer Vision (ICCV)},
  Oct 2017.

\bibitem{saleh2020artist}
Fatemeh Saleh, Sadegh Aliakbarian, Mathieu Salzmann, and Stephen Gould.
\newblock Artist: Autoregressive trajectory inpainting and scoring for
  tracking.
\newblock {\em arXiv preprint arXiv:2004.07482}, 2020.

\bibitem{walker2017pose}
Jacob Walker, Kenneth Marino, Abhinav Gupta, and Martial Hebert.
\newblock The pose knows: Video forecasting by generating pose futures.
\newblock In {\em Computer Vision (ICCV), 2017 IEEE International Conference
  on}, pages 3352--3361. IEEE, 2017.

\bibitem{williams1989learning}
Ronald~J Williams and David Zipser.
\newblock A learning algorithm for continually running fully recurrent neural
  networks.
\newblock {\em Neural computation}, 1(2):270--280, 1989.

\bibitem{yan2018mt}
Xinchen Yan, Akash Rastogi, Ruben Villegas, Kalyan Sunkavalli, Eli Shechtman,
  Sunil Hadap, Ersin Yumer, and Honglak Lee.
\newblock Mt-vae: Learning motion transformations to generate multimodal human
  dynamics.
\newblock In {\em European Conference on Computer Vision}, pages 276--293.
  Springer, 2018.

\bibitem{yang2018diversitysensitive}
Dingdong Yang, Seunghoon Hong, Yunseok Jang, Tiangchen Zhao, and Honglak Lee.
\newblock Diversity-sensitive conditional generative adversarial networks.
\newblock In {\em International Conference on Learning Representations}, 2019.

\bibitem{yuan2020dlow}
Ye Yuan and Kris Kitani.
\newblock Dlow: Diversifying latent flows for diverse human motion prediction.
\newblock In {\em Proceedings of the European Conference on Computer Vision
  (ECCV)}, 2020.

\bibitem{zhang2013actemes}
Weiyu Zhang, Menglong Zhu, and Konstantinos~G Derpanis.
\newblock From actemes to action: A strongly-supervised representation for
  detailed action understanding.
\newblock In {\em Proceedings of the IEEE International Conference on Computer
  Vision}, pages 2248--2255, 2013.

\end{thebibliography}
}
\newpage
\appendix

\section{Pseudo-code for LCP-VAE}
\label{appendix:code}
Here, we provide the forward pass pseudo-code for both \texttt{CS-VAE} and \texttt{LCP-VAE}.

    \begin{algorithm}[!h]
    \small
        \caption{A forward pass of \texttt{CS-VAE}}\label{ovae}
        \begin{algorithmic}[1]
            \Procedure{\texttt{CS-VAE}}{$condition$}               \Comment{Human motion up to time $t$ or source text}
            \State $h_t$ = \texttt{EncodeCondition}($x_t$)        \Comment{Observed motion/source text encoder}
            \State $\mu_c$, $\sigma_c$ = \texttt{CS-VAE.Encode}($h_t$)
             \State \texttt{Sample} $\epsilon\sim\mathcal{N}(0,I)$   \Comment{Sample from standard Gaussian}
            \State $z_c$ = $\mu_c + \sigma_c \odot \epsilon$    \Comment{Reparameterization}
            \State $\hat{h}_t$ = \texttt{CS-VAE.Decode}($z_c$)
            \State $\hat{x}_t$ = \texttt{DecodeCondition}($\hat{h}_t$, \texttt{seed}) 
            \State \textbf{return}  $\hat{x}_t$, $\mu_c$, $\sigma_c$, $h_t$, $z_c$ 
            % \Comment{$h_t$ and $z_c$ condition \texttt{LCP-VAE} encoder and decoder respectively}
            \EndProcedure
        \end{algorithmic}
    \end{algorithm}

    \begin{algorithm}[!h]
    \small
        \caption{A forward pass of \texttt{LCP-VAE}}\label{fmvae}
        \begin{algorithmic}[1]
            \Procedure{\texttt{LCP-VAE}}{$x_T$, $z_c$, $h_t$}               \Comment{Human motion from $t$ to $T$ or target text}
            \If{\texttt{isTraining}}
                \State $h_T$ = \texttt{EncodeData}($x_T$)        \Comment{Future motion/target sentence encoder}
                \State $h_{Tt}$ = \texttt{Concatenate}($h_T$, $h_t$)
                \State $\mu$, $\sigma$ = \texttt{LCP-VAE.Encode}($h_{Tt}$)
                \State $z$ = $\mu + \sigma \odot z_c$   \Comment{Our extended reparameterization}
            \Else
                \State $z$ = $z_c$
            \EndIf
            \State $\hat{h}_T$ = \texttt{LCP-VAE.Decode}($z$)
            \State $\hat{x}_T$ = \texttt{DecodeData}($\hat{h}_T$, \texttt{seed}) 
            
            \State \textbf{return}  $\hat{x}_T$, $\mu$, $\sigma$  
            \EndProcedure
        \end{algorithmic}
    \end{algorithm}

\section{Derivation of LCP-VAE's KL Divergence Loss}
\label{appendix:KL}
In our approach, the model encourages the posterior of \texttt{LCP-VAE} to be close to the one of \texttt{CS-VAE}. In general, the KL divergence between two distributions $P_1$ and $P_2$ is defined as
\begin{align}
    \mathcal{D}_{KL}(P_1 || P_2) = \mathbb{E}_{P_1} \bigg[\log\frac{P_1}{P_2}\bigg]\;.
\end{align}
Let us now consider the case where the distributions are multivariate Gaussians $\mathcal{N}(\mu, \Sigma)$  in $\mathbb{R}^d$, where $\Sigma=\text{diag}(\sigma^2)$, with $\sigma$ and $\mu$ are $d$-dimensional vectors predicted by the encoder network of the VAE. The density function of such a distribution is 
\begin{align}
    p(x) = \frac{1}{(2\pi)^{\frac{d}{2}} det(\Sigma)^{\frac{1}{2}}} exp \bigg(-\frac{1}{2}(x-\mu)^T\Sigma^{-1}(x-\mu)\bigg)\;.
\end{align}{}
Thus, the KL divergence between two multivariate Gaussians is computed as
{\small
\begin{flalign}
    &\mathcal{D}_{KL}(P_1 || P_2) \nonumber\\
    &= \frac{1}{2} \mathbb{E}_{P_1}\bigg[-\log \det \Sigma_1 - (x-\mu_1)^T\Sigma_1^{-1}(x-\mu_1) + \nonumber \\&\log \det\Sigma_2+(x-\mu_2)^T\Sigma_2^{-1}(x-\mu_2)\bigg]
    \nonumber & \\
    & = \frac{1}{2}\log\frac{\det\Sigma_2}{\det\Sigma_1} + \frac{1}{2} \mathbb{E}_{P_1}\bigg[ - (x-\mu_1)^T\Sigma_1^{-1}(x-\mu_1) +\nonumber \\&(x-\mu_2)^T\Sigma_2^{-1}(x-\mu_2)\bigg]
    \nonumber & \\
    & = \frac{1}{2}\log\frac{\det\Sigma_2}{\det\Sigma_1} + \frac{1}{2} \mathbb{E}_{P_1}\bigg[-tr\{\Sigma_1^{-1}(x-\mu_1)(x-\mu_1)^T\} +\nonumber \\& tr\{\Sigma_2^{-1}(x-\mu_2)(x-\mu_2)^T\}\bigg]
    \nonumber & \\
    & = \frac{1}{2}\log\frac{\det\Sigma_2}{\det\Sigma_1} + \frac{1}{2} \mathbb{E}_{P_1}\bigg[-tr\{\Sigma_1^{-1}\Sigma_1\} +\nonumber \\&tr\{\Sigma_2^{-1}(xx^T-2x\mu^T_2+\mu_2\mu_2^T)\}\bigg]
    \nonumber & \\
    & = \frac{1}{2}\log\frac{\det\Sigma_2}{\det\Sigma_1} - \frac{1}{2}d + \frac{1}{2}tr\{\Sigma_2^{-1}(\Sigma_1+\mu_1\mu_1^T-\nonumber \\&2\mu_2\mu_1^T+\mu_2\mu_2^T)\}
    \nonumber & \\
    & = \frac{1}{2}\Bigg[\log\frac{\det\Sigma_2}{\det\Sigma_1} - d + tr\{\Sigma_2^{-1}\Sigma_1\}+tr\{\mu_1^T\Sigma_2^{-1}\mu_1 - \nonumber \\&2\mu_1^T\Sigma_2^{-1}\mu_2 + \mu_2^T\Sigma_2^{-1}\mu_2\}\Bigg]
    \nonumber & \\
    & = \frac{1}{2}\Bigg[\log\frac{|\Sigma_2|}{|\Sigma_1|}-d+tr\{\Sigma_2^{-1}\Sigma_1\} +\nonumber \\& (\mu_2 - \mu_1)^T\Sigma_2^{-1}(\mu_2 - \mu_1)\Bigg]\;.
    \label{eq:kl}
\end{flalign}
}
where $tr\{\cdot\}$ denotes the trace operator. In Eq.~\ref{eq:kl}, the covariance matrix $\Sigma_1$ and mean $\mu_1$ correspond to distribution $P_1$ and the covariance matrix $\Sigma_2$ and mean $\mu_2$ correspond to distribution $P_2$. 

Given 
this result, we can then compute the KL divergence of the \texttt{LCP-VAE} and the posterior distribution with mean  $\mu + \sigma\odot\mu_c$ and covariance matrix $\text{diag}((\sigma\odot\sigma_c)^2)$. Let $\Sigma=\text{diag}(\sigma^2)$,  $\Sigma_c=\text{diag}(\sigma_c^2)$, and $d$ be the dimensionality of the latent space. The loss in Eq. 7 of the main paper can then be written as
\begin{align}
    \mathcal{L}_{prior}^{\texttt{LCP-VAE}} = -\frac{1}{2}\Big[\log\frac{|\Sigma_c|}{|\Sigma_c||\Sigma|}-d+
    tr\{\Sigma_c^{-1}\Sigma_c\Sigma\}+ \nonumber \\(\mu_c-(\mu+\Sigma\mu_c))^T\Sigma_c^{-1}(\mu_c-(\mu+\Sigma\mu_c))\Big]\;.
\end{align}

Since $\Sigma_c^{-1}\Sigma_c=I$, $|\Sigma_c|$ will be cancelled out in the $\log$ term, which yields
\begin{align}
    \mathcal{L}_{prior}^{\texttt{LCP-VAE}} = & -\frac{1}{2}\Big[\log\frac{1}{|\Sigma|}-d+
    tr\{\Sigma\} + \nonumber \\ & (\mu_c-(\mu+\Sigma\mu_c))^T\Sigma_c^{-1}(\mu_c-(\mu+\Sigma\mu_c))\Big]\;. 
\end{align}

\section{LCP-VAE Architecture}
\label{appendix:motionarch}
Our motion prediction model follows the architecture depicted in Fig. 3 of the main paper. Below, we describe the architecture of each component in our model. Note that human poses, consisting of 32 joints in case of the Human3.6M dataset, are represented in 4D quaternion space. Thus, each pose at each time-step is represented with a vector of size $1\times 128$. All the tensor sizes described below ignore the mini-batch dimension for simplicity. 

The \textbf{observed motion encoder}, or \texttt{CS-VAE} motion encoder, is a single layer GRU~\cite{chung2014empirical} network with 1024 hidden units. If the observation sequence has length $T_{obs}$, the observed motion encoder maps $T_{obs}\times 128$ into a single hidden representation of size $1\times 1024$, i.e., the hidden state of the last time-step. This hidden state, $h_t$, acts as the condition to the \texttt{LCP-VAE} encoder and the direct input to the \texttt{CS-VAE} encoder.

\textbf{\texttt{CS-VAE}}, similarly to any variational autoencoder, has an encoder and a decoder. The \texttt{CS-VAE} encoder is a fully-connected network with ReLU non-linearities, mapping the hidden state of the motion encoder, i.e., $h_t$, to an embedding of size $1\times 512$. Then, to generate the mean and standard deviation vectors, we use two fully connected branches. They map the embedding of size $1\times 512$ to a mean vector  of size $1\times 128$ and a standard deviation vector of size $1\times 128$, where 128 is the length of the latent variable. Note that we apply a ReLU non-linearity to the vector of standard deviations to ensure that it is non-negative. We then use the reparameterization trick~\cite{kingma2013auto} to sample a latent variable of size $1\times 128$. The \texttt{CS-VAE} decoder consists of multiple fully-connected layers, mapping the latent variable to a variable of size $1\times 1024$, acting as the initial hidden state of the observed motion decoder. Note that we apply a Tanh non-linearity to the generated hidden state to mimic the properties of a GRU hidden state.

The \textbf{observed motion decoder}, or \texttt{CS-VAE} motion decoder, is similar to its motion encoder, except for the fact that it reconstructs the motion autoregressively. Additionally, it is initialized with the reconstructed hidden state, i.e., the output of the \texttt{CS-VAE} decoder. The output of each GRU cell at each time-step is then fed to a fully-connected layer, mapping the GRU output to a vector of size $1\times 128$, which represents a human pose with 32 joints in 4D quaternion space. To decode the motions, we use a teacher forcing technique~\cite{williams1989learning} during training. At each time-step, the network chooses with probability $P_{tf}$ whether to use its own output at the previous time-step or the ground-truth pose as input. We initialize $P_{tf}=1$, and decrease it linearly at each training epoch such that, after a certain number of epochs, the model becomes completely autoregressive, i.e., uses only its own output as input to the next time-step. Note that, at test time, the motions are generated completely autoregressively, i.e., with $P_{tf}=0$.

Note that the future motion encoder and decoder have exactly the same architectures as the observed motion ones. The only difference is their input, where the future motion is represented by poses from $T_{obs}$ to $T_{end}$ in a sequence. In the following, we describe the architecture of \texttt{LCP-VAE} for motion prediction.

\textbf{\texttt{LCP-VAE}} is a conditional variational encoder. Its encoder's input is a representation of future motion, i.e., the last hidden state of the future motion encoder, $h_T$, conditioned on $h_t$. The conditioning is done by concatenation, thus the input to the encoder is a representation of size $1\times 2048$. The \texttt{LCP-VAE} encoder, similarly to the \texttt{CS-VAE} encoder, maps its input representation to an embedding of size $1\times 512$. Then, to generate the mean and standard deviation vectors, we use two fully connected branches, mapping the embedding of size $1\times 512$ to a mean vector of size $1\times 128$ and a standard deviation vector of size $1\times 128$, where 128 is the length of the latent variable. Note that we apply a ReLU non-linearity to the vector of standard deviations to ensure that it is non-negative. To sample the latent variable, we use our extended reparameterization trick, explained in the main paper. This unifies the conditioning and sampling of the latent variable. Then, similarly to \texttt{CS-VAE}, the latent variable is fed to the \texttt{LCP-VAE} decoder, which is a fully connected network that maps the latent representation of size $1\times 128$ to a reconstructed hidden state of size $1\times 1024$ for future motion prediction. Note that we apply a Tanh non-linearity to the generated hidden state to mimic the properties of a GRU hidden state.

To train our model, we use Adam optimizer with learning rate of 0.005 and mini-batch size of 128. We train our model for 100 epochs on a single NVIDIA GTX 2080Ti.

\section{Results on the Penn Action Dataset}
\label{appendix:penn}
As a complementary experiment, we evaluate our approach on the Penn Action dataset, which contains 2326 sequences of 15 different actions, where for each person, 13 joints are annotated in 2D space. Most sequences have less than 50 frames and the task is to generate the next 35 frames given the first 15. Results are provided in Table~\ref{tab:penn}, where we compare our approach with the deterministic autoregressive (AR) counterpart. Note that the upper bound for the Context metric is 0.74, i.e., the classification performance given the Penn Action ground-truth motions.

\begin{table}[!h]
    \centering
    \small
    \caption{Quantitative evaluation on the Penn Action dataset. Note that a diversity of 1.21 is reasonably high for normalized 2D joint positions, i.e., values between 0 and 1, normalized with the width and the height of the image.}
    \scalebox{0.9}{
    \begin{tabular}{l c c c c}
    \toprule
     & Test MSE (KL) & Diversity & Quality & Context\\
      Method & \scriptsize{(Reconstructed)} & \scriptsize{(Sampled)} & \scriptsize{(Sampled)} & \scriptsize{(Sampled)} \\
     \midrule
        \texttt{LCP-VAE} &  0.034 (6.07) & 1.21 & 0.46 & 0.70\\
        AR Counterpart & 0.048 (N/A) & 0.00 & 0.46 & 0.51\\
        \bottomrule
    \end{tabular}
    }
    \label{tab:penn}
\end{table}{}

\section{Evaluating Sampling Quality}
In Table~\ref{tab:deterministic}, we compare our approach with the state-of-the-art deterministic motion prediction models~\cite{martinez2017human,jain2016structural,gui2018few,fragkiadaki2015recurrent,gui2018adversarial} using the MAE metric in Euler space. 
To have a fair comparison, we generate one motion per observation by setting the latent variable to the distribution mode, i.e., $z=\mu_c$. This allows us to generate a plausible motion without having access to the ground truth. To compare against the deterministic baselines, we follow the standard setting, and thus use 50 frames (i.e., 2sec) as observation to generate the next 25 frames (i.e., 1sec). Surprisingly, despite having a very simple motion decoder architecture (one-layer GRU network) with a very simple reconstruction loss function (MSE), this motion-from-mode strategy yields results that are competitive with those of the baselines that use sophisticated architectures and advanced loss functions. We argue that learning a good, context-preserving latent representation of human motion is the contributing factor to the success of our approach. This, however, could be used in conjunction with sophisticated motion decoders and reconstruction losses, which we leave for future research.

\begin{table}[t]
    \centering
    \scriptsize
    \caption{Comparison with the state-of-the-art deterministic models for 4 actions of Human3.6M. Note that in our approach, we use $z=\mu_c$ to generate a single motion.}
    \scalebox{0.93}{
    \begin{tabular}{l @{ }@{ } c @{ }@{ } c @{ }@{ } c @{ }@{ } c @{ }@{ } c @{ }@{ } c @{ }@{ } @{ }@{ } c @{ }@{ } c @{ }@{ } c @{ }@{ } c @{ }@{ } c @{ }@{ } c}
\toprule
    & \multicolumn{6}{c}{Walking} & \multicolumn{6}{c}{Eating} \\
    \midrule
    Method & 80  & 160 & 320 & 400 & 560 & 1000  & 80  & 160 & 320 & 400 & 560 & 1000  \\
    \midrule
    Zero Velocity & 
    0.39 & 0.86 & 0.99 & 1.15 & 1.35 & 1.32 & 
    0.27 & 0.48 & 0.73 & 0.86 & 1.04 & 1.38 \\
    
    LSTM-3LR~\cite{fragkiadaki2015recurrent} & 
    1.18 & 1.50 & 1.67 & 1.76 & 1.81 & 2.20  & 
    1.36 & 1.79 & 2.29 & 2.42 & 2.49 & 2.82 \\
    
    SRNN~\cite{jain2016structural} & 
    1.08 & 1.34 & 1.60 & 1.80 & 1.90 & 2.13 & 
    1.35 & 1.71 & 2.12 & 2.21 & 2.28 & 2.58 \\
    
    DAE-LSTM~\cite{ghosh2017learning} & 
    1.00 & 1.11 & 1.39 & 1.48 & 1.55 & 1.39 & 
    1.31 & 1.49 & 1.86 & 1.89 & 1.76 & 2.01 \\ 
    
    GRU~\cite{martinez2017human} & 
    0.28 & 0.49 & 0.72 & 0.81 & 0.93 & 1.03 & 
    0.23 & 0.39 & 0.62 & 0.76 & 0.95 & 1.08 \\

    AGED~\cite{gui2018adversarial} & 
    0.22 & 0.36 & 0.55 & 0.67 & 0.78 & 0.91 & 
    0.17 & 0.28 & 0.51 & 0.64 & 0.86 & 0.93 \\

    DCT-GCN~\cite{wei2019motion} & 
    \textbf{0.18} & \textbf{0.31} & 0.49 & 0.56 & 0.65 & \textbf{0.67} & 
    \textbf{0.16} & 0.29 & 0.50 & 0.62 & 0.76 & 1.12 \\
    
    \texttt{LCP-VAE} & 
    0.20 & 0.34 & \textbf{0.48} & \textbf{0.53} & \textbf{0.57} & 0.71 & 
    0.20 & \textbf{0.26} & \textbf{0.44} & \textbf{0.52} & \textbf{0.61} & \textbf{0.92} \\
    
    %%%%%%%%%%%%%%%%%%%%%%
    \midrule
    & \multicolumn{6}{c}{Smoking} & \multicolumn{6}{c}{Discussion}\\
    \midrule
    Method & 80  & 160 & 320 & 400 & 560 & 1000  & 80  & 160 & 320 & 400 & 560 & 1000  \\
    \midrule
    Zero Velocity & 
    0.26 & 0.48 & 0.97 & 0.95 & 1.02 & 1.69 & 
    0.31 & 0.67 & 0.94 & 1.04 & 1.41 & 1.96 \\
    
    LSTM-3LR~\cite{fragkiadaki2015recurrent} & 
    2.05 & 2.34 & 3.10 & 3.18 & 3.24 & 3.42  & 
    2.25 & 2.33 & 2.45 & 2.46 & 2.48 & 2.93 \\
    
    SRNN~\cite{jain2016structural} & 
    1.90 & 2.30 & 2.90 & 3.10 & 3.21 & 3.23 & 
    1.67 & 2.03 & 2.20 & 2.31 & 2.39 & 2.43 \\
    
    DAE-LSTM~\cite{ghosh2017learning} & 
    0.92 & 1.03 & 1.15 & 1.25 & 1.38 & 1.77 & 
    1.11 & 1.20 & 1.38 & 1.42 & 1.53 & 1.73 \\
    
    GRU~\cite{martinez2017human} & 
    0.33 & 0.61 & 1.05 & 1.15 & 1.25 & 1.50 & 
    0.31 & 0.68 & 1.01 & 1.09 & 1.43 & 1.69 \\

    AGED~\cite{gui2018adversarial} & 
    0.27 & 0.43 & 0.82 & 0.84 & 1.06 & 1.21 & 
    0.27 & 0.56 & \textbf{0.76} & 0.83 & 1.25 & 1.30 \\

    DCT-GCN~\cite{wei2019motion} & 
    0.22 & \textbf{0.41} & 0.86 & 0.80 & 0.87 & 1.57 & 
    \textbf{0.20} & \textbf{0.51} & 0.77 & 0.85 & 1.33 & 1.70 \\
    
    \texttt{LCP-VAE} & 
    \textbf{0.21} & 0.43 & \textbf{0.79} & \textbf{0.79} & \textbf{0.77} & \textbf{1.15} & 
    0.22 & 0.55 & 0.79 & \textbf{0.81} & \textbf{1.05} & \textbf{1.28}\\
    
\bottomrule
    \end{tabular}
    }
\label{tab:deterministic}
\end{table}

\section{Additional Qualitative Results}
\label{appendix:qualitativemotion}
\begin{figure}[!h]
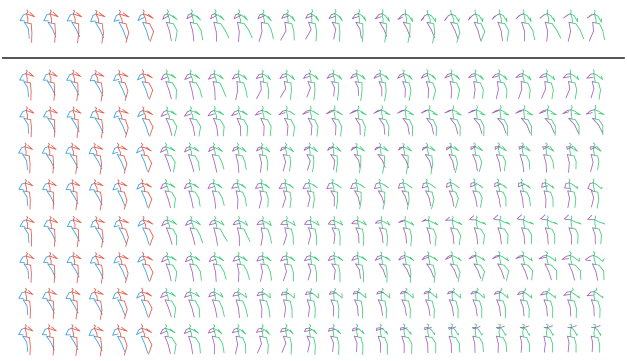

    \centering
    \begin{tabular}{c}
        \toprule
        \includegraphics[width=.45\textwidth]{latex/figures/collage_gt.png}\\
        \midrule
        \includegraphics[width=.45\textwidth]{latex/figures/collage_h36m.png}\\
    \bottomrule
    \end{tabular}
    
    \caption{Qualitative evaluation of the diversity in human motion. The first row illustrates the ground-truth motion. The first six poses of each row depict the observation (the condition) and the rest are sampled from our model. Each row is a randomly sampled motion (not cherry picked). As can be seen, all sampled motions are natural, with a smooth transition from the observed to the generated ones. The diversity increases as we increase the sequence length.}
    \label{fig:qualitative_h36m1}
    % \vspace{20pt}
\end{figure}

\begin{figure}[!h]
    \centering
    \begin{tabular}{c}
    \toprule
    \includegraphics[width=.45\textwidth]{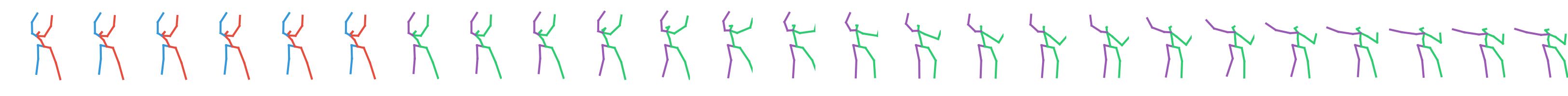}\\
    \midrule
    \includegraphics[width=.45\textwidth]{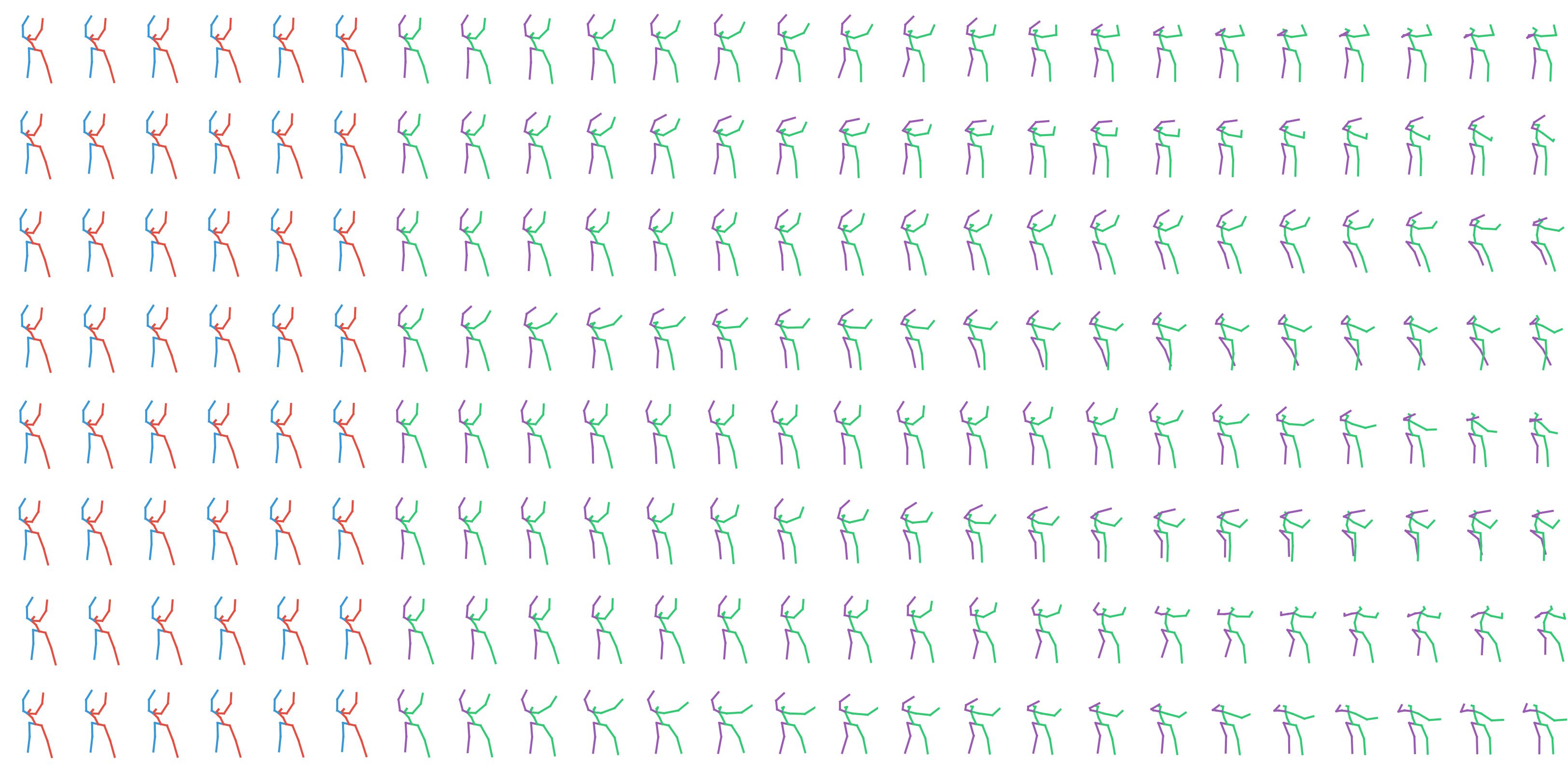}\\
    \bottomrule
    \end{tabular}
    \caption{Additional qualitative evaluation of the diversity in human motion. }
    \label{fig:qualitative_h36m2}
    % \vspace{20pt}
\end{figure}

\begin{figure}[!h]
    \centering
    \begin{tabular}{c}
    \toprule
    \includegraphics[width=.45\textwidth]{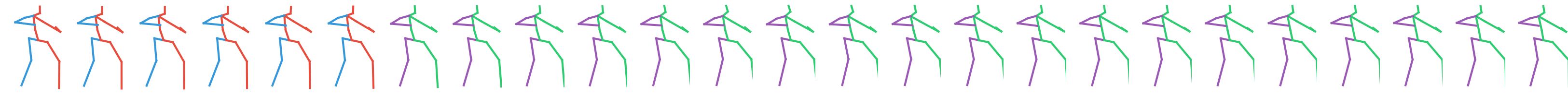}\\
    \midrule
    \includegraphics[width=.45\textwidth]{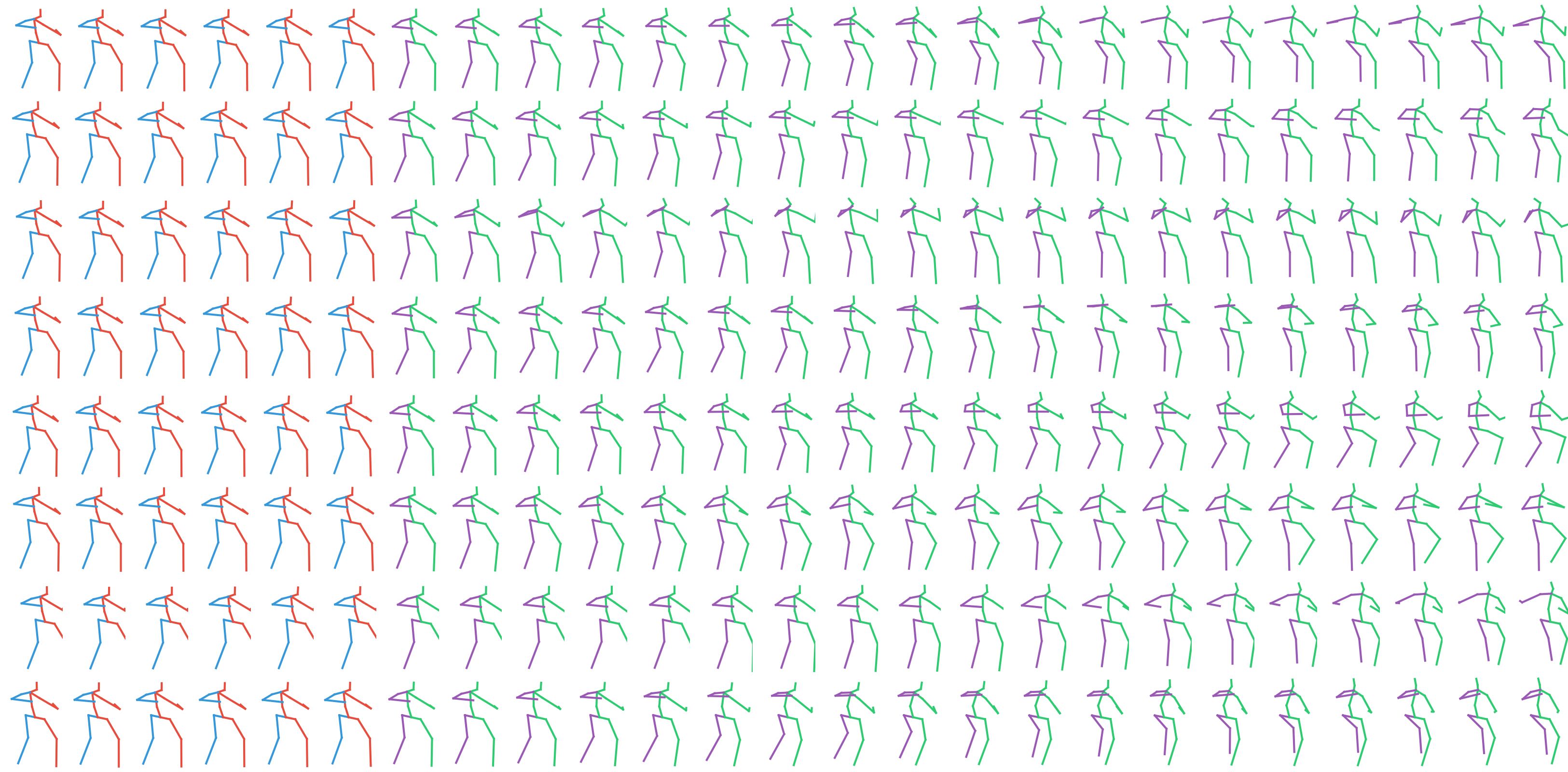}\\
    \bottomrule
    \end{tabular}
    \caption{Additional qualitative evaluation of the diversity in human motion. }
    \label{fig:qualitative_h36m3}
    % \vspace{20pt}
\end{figure}

\begin{figure}[!h]
    \centering
    \begin{tabular}{c}
    \toprule
    \includegraphics[width=.45\textwidth]{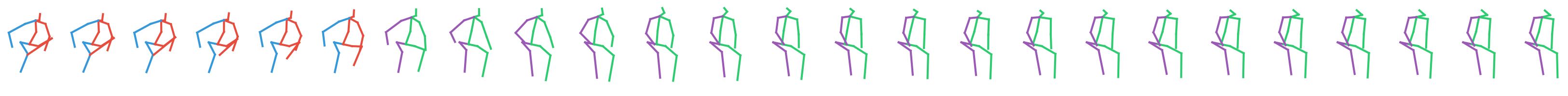}\\
    \midrule
    \includegraphics[width=.45\textwidth]{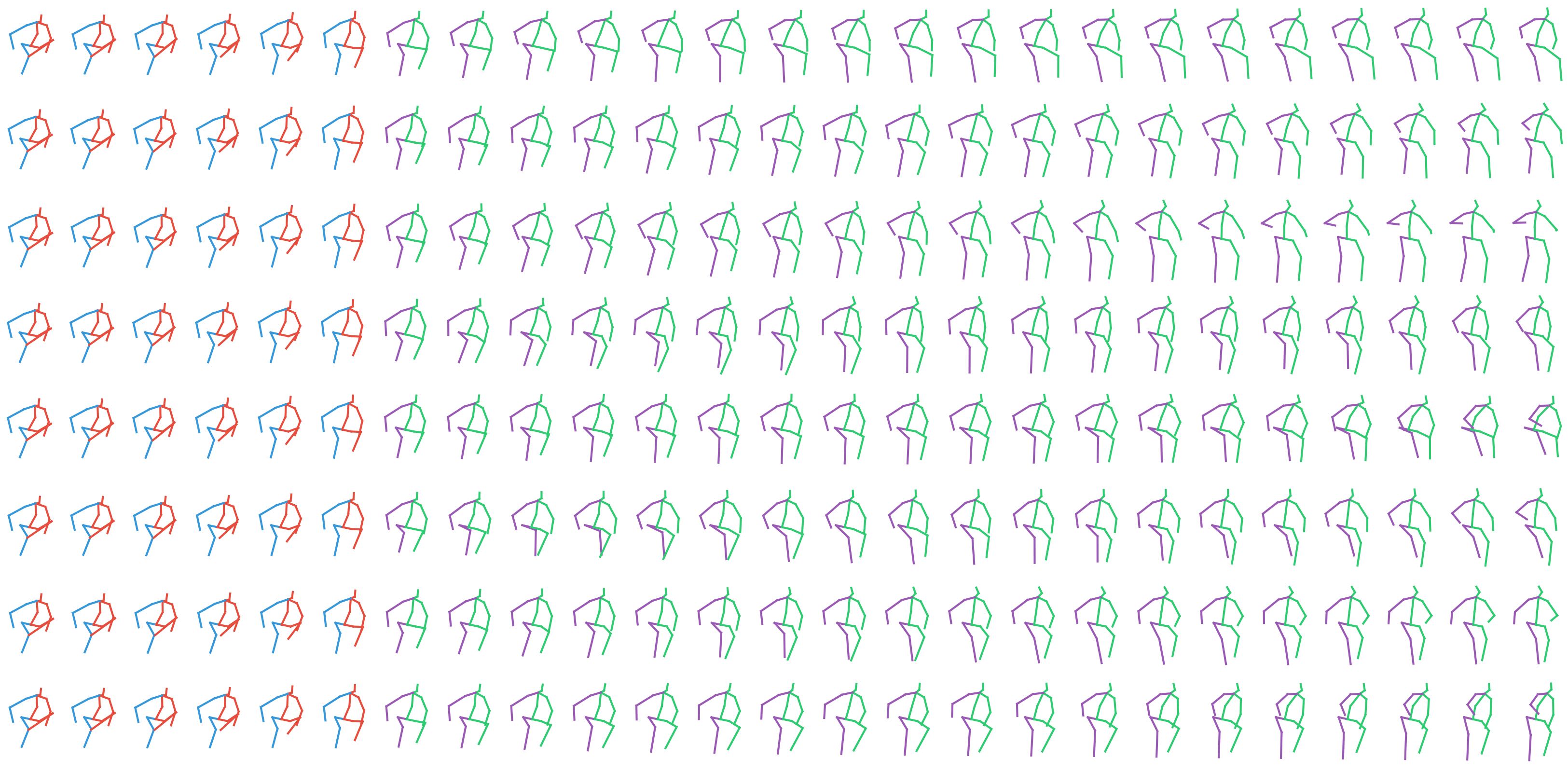}\\
    \bottomrule
    \end{tabular}
    \caption{Additional qualitative evaluation of the diversity in human motion. }
    \label{fig:qualitative_h36m4}
    % \vspace{20pt}
\end{figure}

\begin{figure}[!h]
    \centering
    \begin{tabular}{c}
    \toprule
    \includegraphics[width=.45\textwidth]{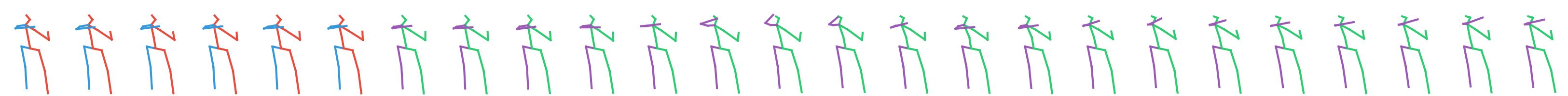}\\
    \midrule
    \includegraphics[width=.45\textwidth]{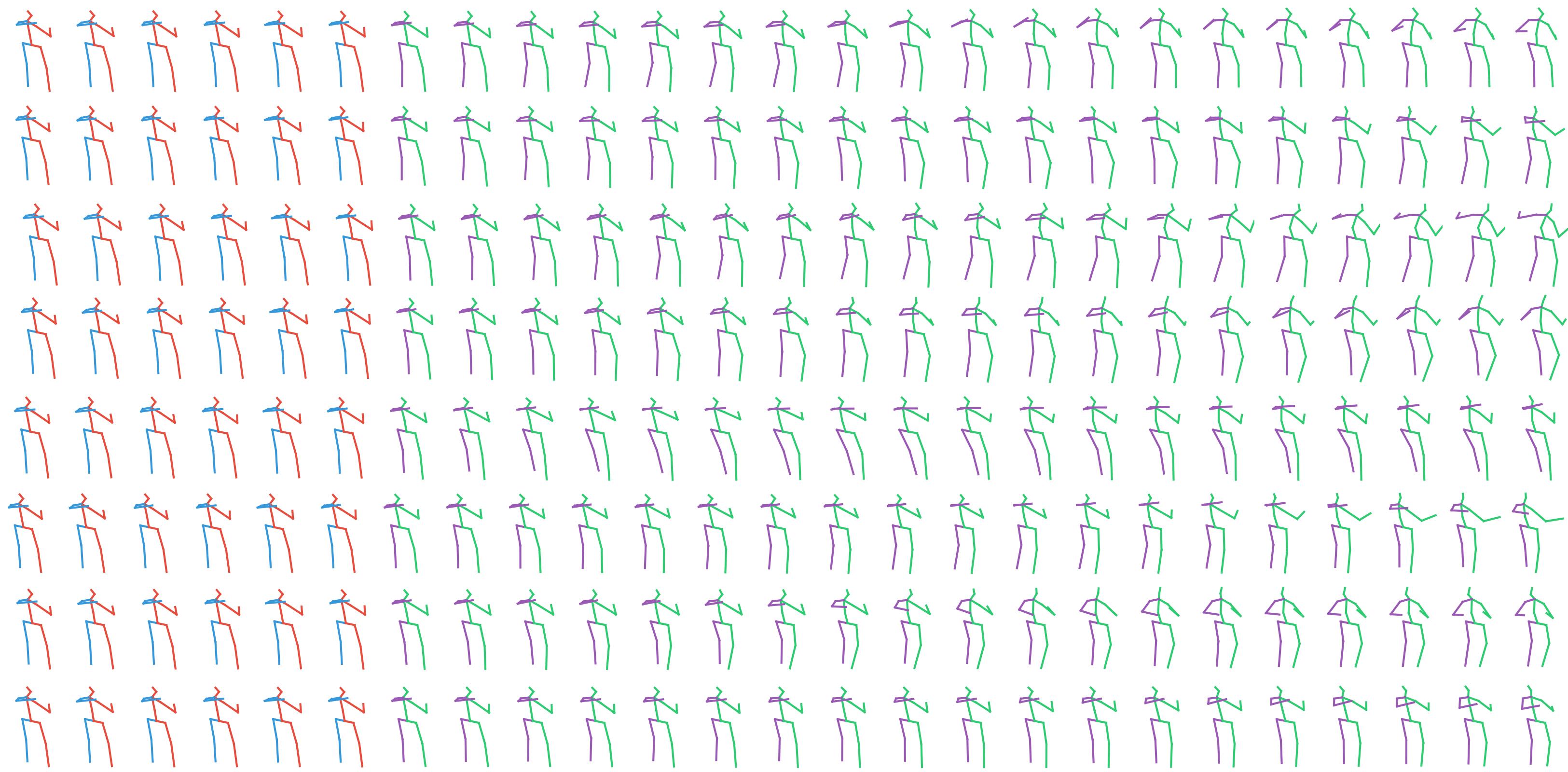}\\
    \bottomrule
    \end{tabular}
    \caption{Additional qualitative evaluation of the diversity in human motion. }
    \label{fig:qualitative_h36m5}
    % \vspace{20pt}
\end{figure}

\begin{figure}[!h]
    \centering
    \begin{tabular}{c}
    \toprule
    \includegraphics[width=.45\textwidth]{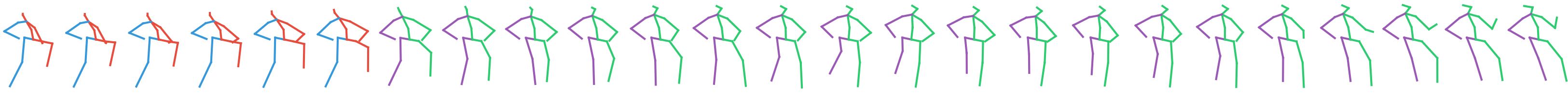}\\
    \midrule
    \includegraphics[width=.45\textwidth]{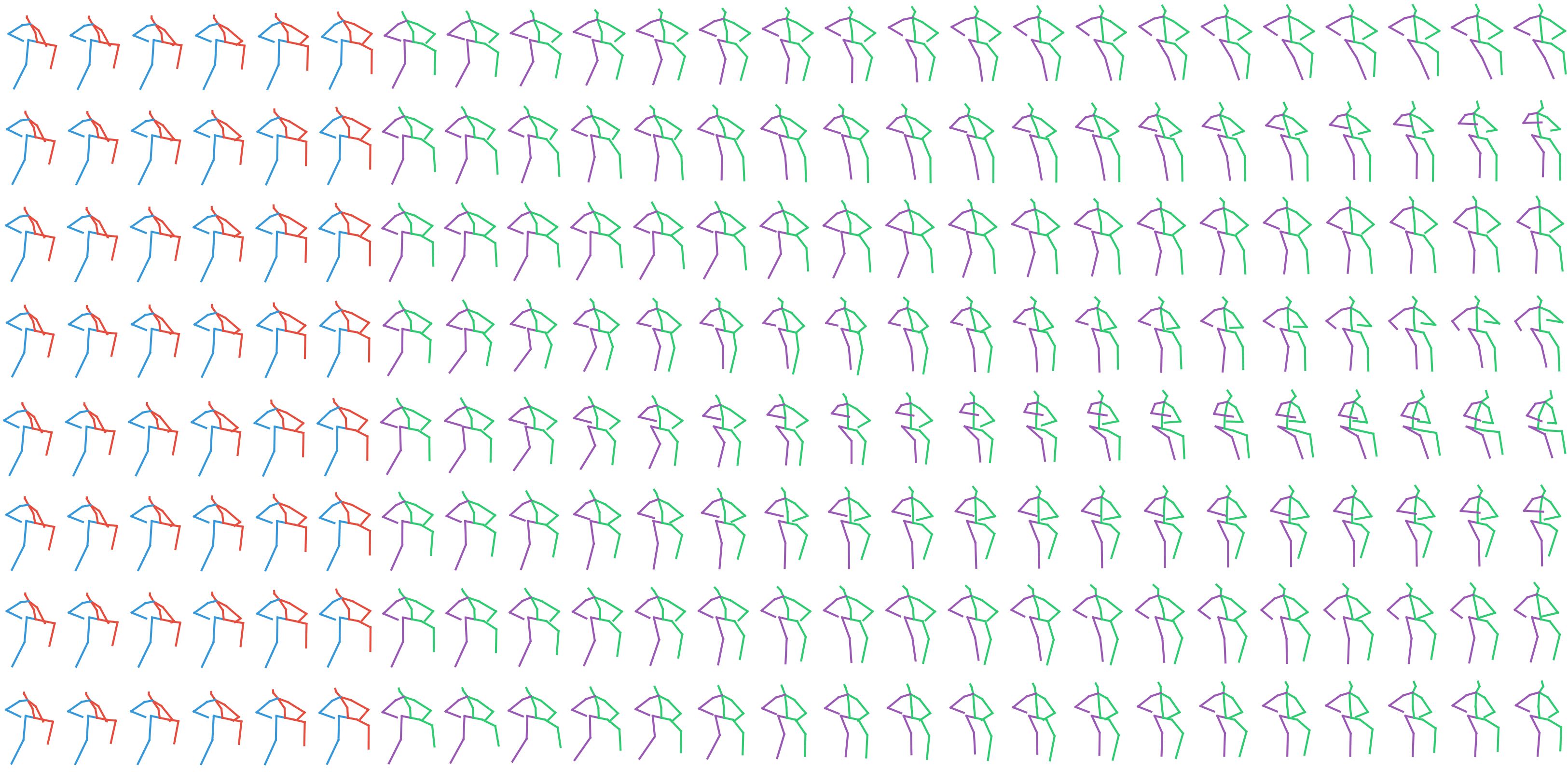}\\
    \bottomrule
    \end{tabular}
    \caption{Additional qualitative evaluation of the diversity in human motion. }
    \label{fig:qualitative_h36m6}
    % \vspace{20pt}
\end{figure}

Here, we provide qualitative results on diverse human motion prediction on the Human3.6M dataset. As can be seen in Figures~\ref{fig:qualitative_h36m1} to~\ref{fig:qualitative_h36m6}, the motions generated by our approach are diverse and natural, and mostly within the context of the observed motion. We also provide qualitative results as a video in a separate file.

{}

{}

{}

{}

{}

{}

\end{document}